\documentclass[10pt,twocolumn,letterpaper]{article}

\usepackage{iccv}              % To produce the CAMERA-READY version
\usepackage{url}
\usepackage{booktabs} 
\usepackage{amsfonts}
\definecolor{Gray}{gray}{0.9}
\usepackage{graphicx}
\usepackage{colortbl}
\usepackage{multirow}
\usepackage{multicol}
\usepackage{bm}
\usepackage{xcolor}    
\usepackage{subcaption}
\usepackage{diagbox}
\usepackage{pifont}
\usepackage{amsmath,mathtools}
\usepackage{amssymb}
\usepackage{animate}
\usepackage{setspace}
\usepackage[accsupp]{axessibility}
\usepackage[inline]{enumitem}
\usepackage{tabularx}
\usepackage{xspace}
\usepackage{interval}
\usepackage{siunitx}
\usepackage{epigraph}
\newcommand{\blank}[1]{\hspace*{#1}}
\newcommand{\system}{\textit{Reducio}\xspace}
\newcommand{\ra}[1]{\renewcommand{\arraystretch}{#1}}

\definecolor{iccvblue}{rgb}{0.21,0.49,0.74}
\definecolor{lightblue}{rgb}{0.93, 0.96, 1.0}
\usepackage[pagebackref,breaklinks,colorlinks,allcolors=iccvblue]{hyperref}

\def\paperID{7412} % *** Enter the Paper ID here
\def\confName{ICCV}
\def\confYear{2025}

\title{
\textit{REDUCIO!} Generating 1K Video within 16 Seconds using \\ Extremely  Compressed Motion Latents
}

\author{Rui Tian$^{1,2}$ \ \ \
Qi Dai$^{3}$\footnotemark[1] \ \ \
Jianmin Bao$^{3}$ \ \ \
Kai Qiu$^3$ \ \ \
Yifan Yang$^3$ \\ 
Chong Luo$^3$ \ \ \
Zuxuan Wu$^{1,2}$\footnotemark[1] \ \ \
Yu-Gang Jiang$^{1,2}$ \\
{$^1$Institute of Trustworthy Embodied AI, Fudan University} \\
{$^2$Shanghai Collaborative Innovation Center of Intelligent Visual Computing} {$^3$Microsoft Research} \\
}

\begin{document}
\twocolumn[{
\maketitle
\vspace{-2.2em}
\renewcommand\twocolumn[1][]{#1}

% \begin{center}
%     \centering
%     \begin{minipage}[b]{0.3884\linewidth}
%         \setstretch{0.6}
%         \animategraphics[width=\linewidth,autoplay=True,keepaspectratio]{8}{figs/vsample_1/00}{00}{15}\\
%         % \vspace{-0.5cm}
%          \textit{\footnotesize Impressionist style, a yellow rubber duck floating on the waves at sunset.}
%     \end{minipage}%
%     \hspace{0.05cm} % Adjust horizontal space between the animations
%     \begin{minipage}[b]{0.6\linewidth}
%         \setstretch{0.6}
%         \animategraphics[width=\linewidth,autoplay=True,keepaspectratio]{8}{figs/vsample_2/00}{00}{15}\\
%         \textit{\footnotesize An aerial shot of a lighthouse standing tall on a rocky cliff, its beacon cutting through the early dawn, waves crashing against the rocks below.}
%     \end{minipage}
%     \vspace{0.15cm}
%      \begin{minipage}{0.998\linewidth}
%         \setstretch{0.6}
%         \includegraphics[width=\linewidth, trim=0 40 205 0, clip]{figs/fig_cover.pdf}\\
%         % \vspace{-0.1cm}
%         \centering
%         \textit{\footnotesize A wizard waves his wand.}
%     \end{minipage}
%     \vspace{-0.4cm}
%     \captionof{figure}{Examples of generated high-resolution videos. {\color{magenta} \emph{Click} the top row images to play the video clips. Best viewed with Adobe Reader.}}
%     \label{fig:cover}
% \end{center}
% }]

\begin{center}
    \centering
    \vspace{-0.15cm}
    % \begin{minipage}[b]{0.4942\linewidth}
    % \setstretch{0.6}\animategraphics[width=\linewidth,autoplay=True,keepaspectratio]{8}{figs/vsample_4/00}{00}{15}\\
    %      \textit{\footnotesize Impressionist style, a yellow rubber duck floating on the waves at sunset. \quad\qquad}
    % \end{minipage}%
    % \hspace{0.05cm} % Adjust horizontal space between the animations
    % \begin{minipage}[b]{0.4942\linewidth}
    %     \setstretch{0.6}
    %     \animategraphics[width=\linewidth,autoplay=True,keepaspectratio]{8}{figs/vsample_5/00}{00}{15}\\
    %     \textit{\footnotesize \qquad An aerial shot of a lighthouse standing tall on a rocky cliff, its beacon cutting through the early dawn, waves crashing against the rocks below.}
    % \end{minipage}
    \begin{minipage}{0.998\linewidth}
        \setstretch{0.6}
        \includegraphics[width=\linewidth]{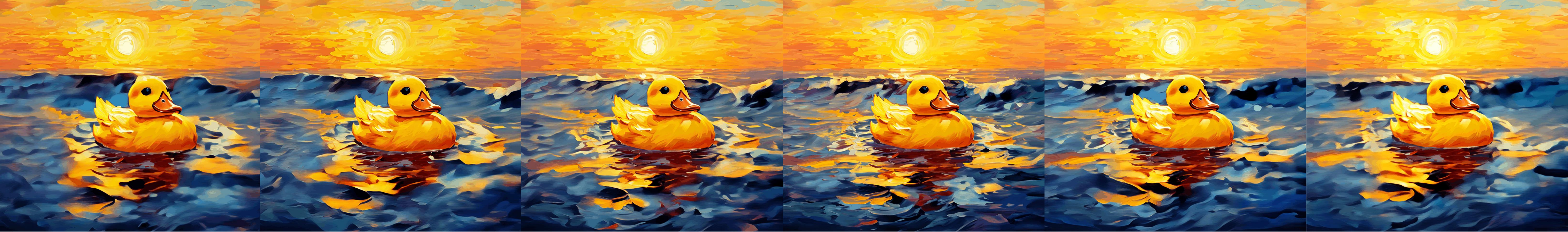}\\
        % \vspace{-0.1cm}
        \centering
        \textit{\footnotesize Impressionist style, a yellow rubber duck floating on the waves at sunset.}
    \end{minipage}
    % \vspace{0.1cm}
    
    \begin{minipage}{0.998\linewidth}
        \setstretch{0.6}
        \includegraphics[width=\linewidth]{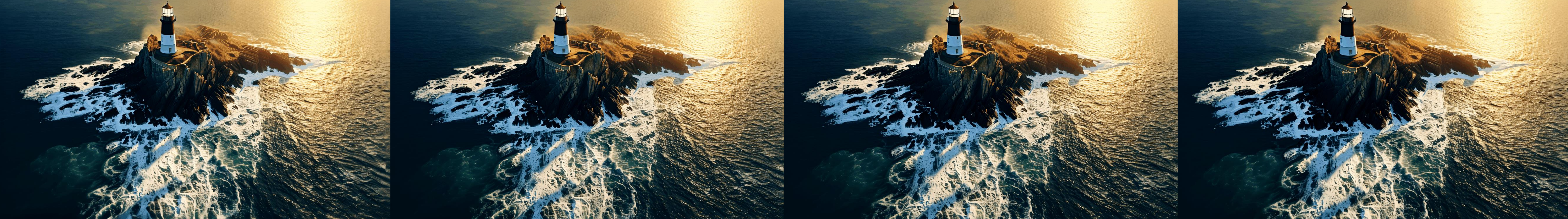}\\
        % \vspace{-0.1cm}
        \centering
        \textit{\footnotesize An aerial shot of a lighthouse standing tall on a rocky cliff, its beacon cutting through the early dawn, waves crashing against the rocks below.}
    \end{minipage}
    % \vspace{0.1cm}
    
     \begin{minipage}{0.998\linewidth}
        \setstretch{0.6}
        \includegraphics[width=\linewidth, trim=0 40 205 0, clip]{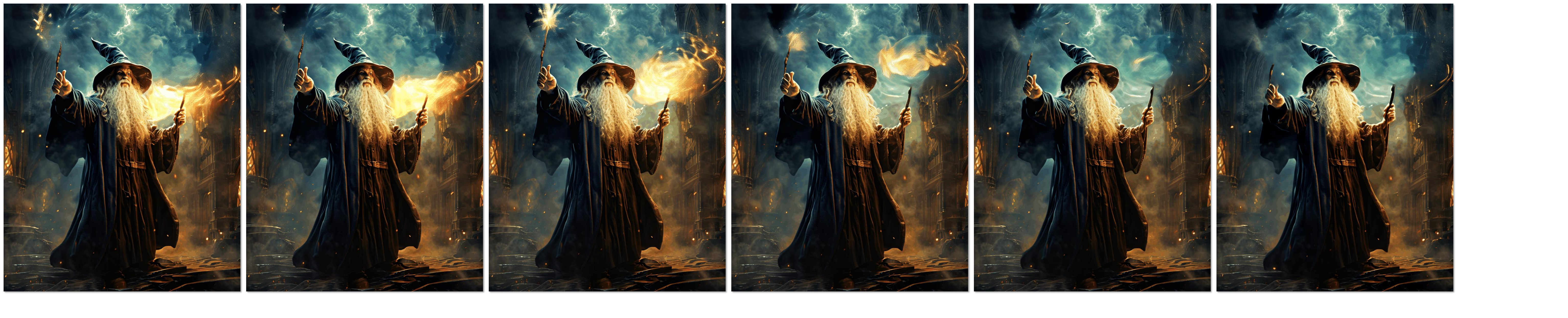}\\
        % \vspace{-0.1cm}
        \centering
        \textit{\footnotesize A wizard waves his wand.}
    \end{minipage}
    \vspace{-0.2cm}
    \captionof{figure}{Examples of generated high-resolution videos. 
    }
    % {\color{magenta} \emph{Click} the top row images to play the video clips. Best viewed with Adobe Reader.}
    \label{fig:cover}
\end{center}

}]

\renewcommand{\thefootnote}{\fnsymbol{footnote}}
\footnotetext[1]{Corresponding authors.}

\begin{abstract}
Commercial video generation models have exhibited realistic, high-fidelity results but are still restricted to limited access.
One crucial obstacle for large-scale applications is the expensive training and inference cost.
In this paper, we argue that videos contain significantly more redundant information than images, allowing them to be encoded with very few motion latents.
Towards this goal, we design an image-conditioned VAE that projects videos into extremely compressed latent space and decode them based on content images. 
This magic Reducio charm enables 64$\times$ reduction of latents compared to a common 2D VAE, without sacrificing the quality.
Building upon Reducio-VAE, we can train diffusion models for high-resolution video generation efficiently. Specifically, 
we adopt a two-stage generation paradigm, first generating a condition image via text-to-image generation, followed by text-image-to-video generation with the proposed Reducio-DiT. 
Extensive experiments show that our model achieves strong performance in evaluation.
More importantly, our method significantly boosts the training and inference efficiency of video LDMs. Reducio-DiT is trained in just 3.2K A100 GPU hours in total and can generate a 16-frame 1024$\times$1024 video clip within 15.5 seconds on a single A100 GPU. Code is available at \url{https://github.com/microsoft/Reducio-VAE}.

\end{abstract}    
\section{Introduction}
\label{sec:intro}
% \epigraph{The straightforward but surprisingly dangerous charm, \system, cause certain things to shrink.}{\textit{Miranda Goshawk \blank{0.2cm}``Harry Potter''~\cite{rowling2019harry}}}
% \vspac

Recent advances in latent diffusion models (LDMs)~\cite{stablediffusion} for video generation have presented inspiring results~\cite{xing2023survey,blattmann2023align,latentshift,blattmann2023stable,videofactory,yang2024cogvideox,bar2024lumiere,chen2023videocrafter1,qing2024hierarchical,weng2024art,tu2025stableanimator,tu2023implicit,tu2024motionfollower,tu2024motioneditor}, showing great potential in various applications.
Commercial models like Sora~\cite{sora}, Runway Gen-3~\cite{gen3}, Movie Gen~\cite{polyak2024movie}, and Kling~\cite{kling} can already generate photorealistic and high-resolution video clips.
However, training and deploying such models are computationally demanding---thousands of GPUs and millions of GPU hours are required for training, and the inference for a one-second clip costs several minutes. 
Such a high cost is becoming a significant obstacle for research and large-scale real-world applications.

Extensive effort has been made to relieve the computational burden by means of efficient computation modules in the backbone~\cite{teng2024dim,xing2024simda}, optimizing diffusion training strategies~\cite{hang2023efficient,wang2024patch}, and adopting few-step sampling methods~\cite{song2023consistency, xu2024ufogen}. However, in this paper, we argue that the very nature of the problem is long neglected. Concretely, most video LDMs stick to the paradigm of text-to-image diffusion, \eg, Stable Diffusion (SD)~\cite{stablediffusion}, and inherently employ the latent space of its pre-trained 2D variational autoencoder (VAE)~\cite{vae,rezende2014stochastic}. This scheme compresses the input videos by $8\times$ in each spatial dimension, which is suitable for images while being much surplus for video.
%They either employ the identical latent space with pre-trained 2D variational autoencoder (VAE)~\cite{}, compressing the input videos by $8\times$ in each spatial dimension.

As videos naturally carry much more redundancy than images, we believe they can be projected into a much more compressed latent space with a VAE based on 3D convolutions. Surprisingly, for videos, a $16\times$ spatial down-sampling factor does not sacrifice much of the reconstruction performance. While the resulting distribution of latent space shifts from those of the pre-trained image LDMs, we can still wield their well-learned spatial prior by factorizing text-to-video generation with a condition frame~\cite{girdhar2023emu,wang2024microcinema,xing2025dynamicrafter,zeng2024make}. In particular, we can obtain a content frame with off-the-shelf text-to-image LDMs and then generate samples with text and image as joint priors into video LDMs.

More importantly, as the image prior delivers enriched spatial information of video contents, factorized video generation offers the potential for further compression. 
We argue that video can be encoded into very few latents that represent motion variables, \ie, motion latents, plus a content image.
%very few latents representing motion variables, plus an image content
%encoded into very few latents representing motion variables by leveraging the content image prior.
Hereby, we introduce the magical \system charm by building a VAE comprised of a 3D encoder that aggressively compresses input videos into a $4096 \times$ down-sampled compact space, and a 3D decoder fused with feature pyramids of the middle frame as the content condition. Notably, the resulting \system-VAE surpasses the common 2D VAE by 5db in PSNR while using a $\mathbf{64\times}$ smaller latent space.

Subsequently, we establish the LDM with a diffusion transformer (DiT)~\cite{peebles2023scalable,chen2023pixart}. Besides using T5 features~\cite{raffel2020exploring} as the text condition, we employ an image semantic encoder as well as a context encoder to provide additional image conditions, which inform the model of the spatial content.
Thanks to the extremely compressed video latent, our diffusion model, namely \system-DiT, thus enjoys both the fast training (inference) speed and high generation quality.
Specifically, \system-DiT achieves 318.5 FVD score on UCF-101, surpassing a bunch of previous work~\cite{blattmann2023align,latentshift,wang2023lavie,zhang2024show,wang2024microcinema} while being much faster.
It can also be easily upscaled to a larger resolution, \eg $1024^2$, with affordable cost.
Our experiments show that \system-DiT can generate a 16-frame video clip of $1024^2$ resolution within 15.5 seconds on a single A100 GPU, and hence achieves a $\mathbf{16.6\times}$ speedup over Lavie~\cite{wang2023lavie}.

In summary, our contributions are as follows: (1) We encode video into very few motion latents plus a content image, where the designed \system-VAE can compress video into a $\mathbf{64\times}$ smaller latent space compared to common 2D VAE.
(2) Based on \system-VAE, a diffusion model called \system-DiT is devised by incorporating multimodal conditions to perform text-image to video generation.
(3) Experiments show that our \system-DiT can significantly speed up the generation process while producing high-quality videos.

\section{Related Work}
\label{sec:rel_work}

\begin{figure*}[ht]
\centering
% \vspace{-0.1in}
\includegraphics[width=\linewidth]{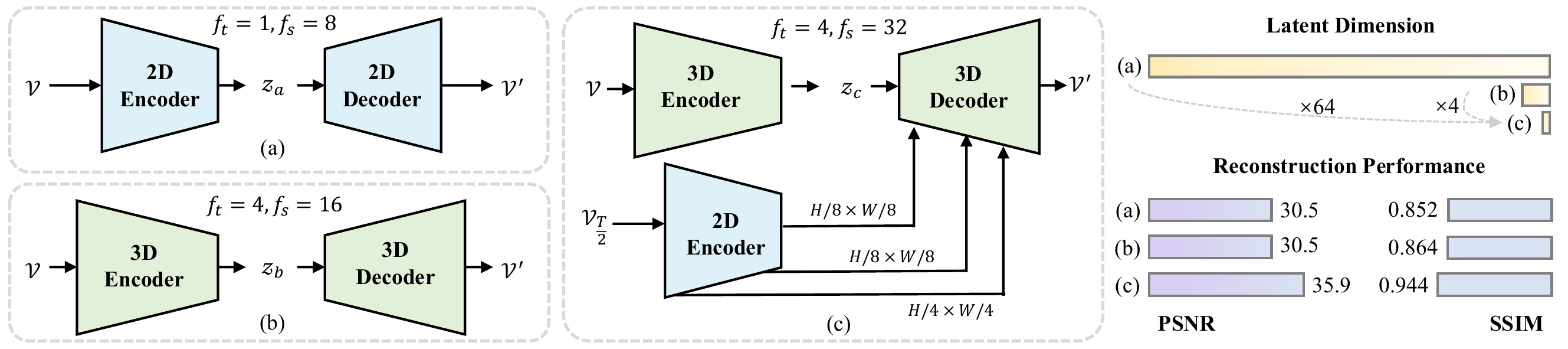}
\vspace{-0.25in}
\caption{Architecture and performance comparisons between vanilla 2D VAE in  (a) SDXL,  (b) 3D VAE and (c) the proposed \system-VAE. \textit{Reducio}-VAE enables 64$\times$ reduction of latent size while achieving much higher reconstruction performance. }
\label{fig:vae_net}
\vspace{-0.15in}
\end{figure*}

\vspace{0.05in}
\noindent \textbf{Video diffusion models} have become the latest craze in the field of text-to-video generation. Due to the huge success of text-to-image LDMs~\cite{ddpm,improvedddpm,song2020score,glide,stablediffusion}, attempts have been made to leverage the powerful UNet-based stable diffusion models pre-trained in 2D space by inserting temporal convolution and attention layers~\cite{feng2024ccedit,xing2024simda,blattmann2023align,blattmann2023stable,videofactory}. Recently, the strong performance of transformers has continued in image generation domain~\cite{wang2025simplear,ma2025token,esser2024scaling,chen2023pixart,flux2024}. Consequently, recent works have focused on extending 2D diffusion transformer (DiT)~\cite{peebles2023scalable} to spatiotemporal domain. 

While the mainstream exploration~\cite{polyak2024movie,zheng2024open,lin2024stiv} follows the paradigm of image diffusion models by using text as the only condition, an emerging stream of works~\cite{girdhar2023emu,wang2024microcinema,zeng2024make} decompose text-to-video diffusion into two steps, \ie, text-to-image and text-image-to-video generation. The factorized framework encourages models to focus on motion modeling instead of challenging spatiotemporal joint modeling. In this work, we build an image-to-video model upon DiT. Thanks to the delicate design of \system-VAE, our diffusion model preserves fine details of image priors in generated videos and hence can be used for more high-demanding real-world applications, \eg, advertisement generation.

\vspace{0.05in}
\noindent \textbf{Efficient diffusion models} have gained increasing attention due to the tremendous computational resources and time required for training and inferring samples, especially for high-resolution images and videos. On one hand, researchers have made attempts to build LDMs with more efficient architectures~\cite{fei2024dimba,teng2024dim,xie2024sana,zhao2024real}. For example, DIM~\cite{teng2024dim} uses Mamba as its backbone to ease the computational burden compared to vanilla transformers. Another line of work leverages efficient diffusion techniques to perform fewer sampling steps during inference~\cite{liu2023instaflow,luhman2021knowledge, song2023consistency, xu2024ufogen,zhang2023adadiff} or to accelerate convergence during training~\cite{hang2023efficient,wang2024patch}. For instance, CausVid~\cite{yin2025slow} applies distribution matching distillation between a causal student video generation model and a bidirectional teacher model.

% # For instance, Stable Diffusion 3~\cite{esser2024scaling} and Flux~\cite{flux2024} employ rectified flow~\cite{} for improved training and inference efficiency.

Our work focuses on designing a more compact latent space~\cite{perniaswurstchen}, which straightforwardly and effectively reduces the overall training and inference costs of LDM. PVDM~\cite{yu2023video} and CMD~\cite{yuefficient} propose projecting 3D video samples into three separate yet compact 2D latent spaces. However, this approach requires training separate diffusion models for each space, which inevitably leads to discrepancies during inference.
Our \system-VAE takes after LaMD~\cite{hu2023lamd} to adopt a factorized, compact motion latent space with the help of a content frame for reconstruction. In contrast, we emphasize aggressive compression of the spatial dimension and suggest that a moderate temporal down-sampling factor provides the best trade-off between performance and efficiency. Concurrent research in the image domain also indicates that using a high spatial down-sampling factor on redundant visual content~\cite{tang2024hart,xie2024sana}, such as high-resolution images, significantly improves efficiency. We propose that it is crucial to recognize the importance of spatial redundancy in the video latent space.
% In the meantime, a branch of concurrent research in image domain also suggests that using a high spatial down-sampling factor on redundant visual contents~\cite{tang2024hart,xie2024sana}, \ie, high-resolution images, yields much-improved efficiency. We propose that we initialize to unveil the importance of spatial redundancy in video latent space.
\section{Reducio}

\begin{figure*}[t]
\centering
% \vspace{-0.1in}
\includegraphics[width=0.85\linewidth]{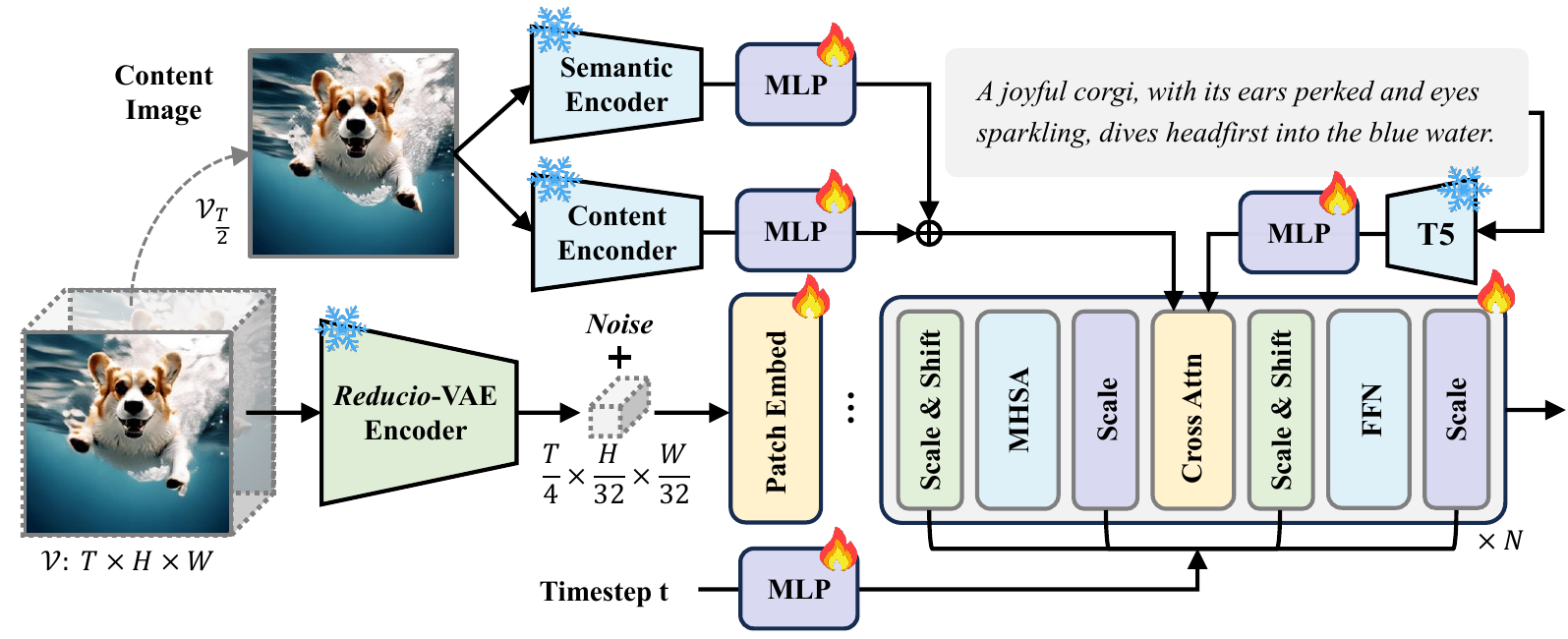}
\vspace{-0.1in}
\caption{The overview of \textit{Reducio}-DiT training framework. Compared to the vanilla DiT, \textit{Reducio}-DiT adopts an additional image-condition module to inject the semantics and content of the keyframe via cross-attention. }
\label{fig:dit_net}
\vspace{-0.15in}
\end{figure*}

\subsection{\textbf{\system} Video Autoencoder}
\vspace{0.05in}
\noindent \textbf{Variational autoencoder} (VAE)~\cite{vae} projects input images or videos into a compressed latent space obeying a certain distribution and maps the latent sampled from the obtained distribution into RGB space with a decoder. Latent diffusion models~\cite{stablediffusion} take advantage of the down-sampled latent space of VAE to support generation with improved efficiency. In this paper, we follow the common practice and build our video autoencoder upon VAE. 

We denote the temporal down-sampling factor of a VAE as $f_t$ and the spatial one as $f_s$. Given an input video $\mathcal{V}$ with the shape of $3 \times T \times  H \times W$, the compressed latent $z_{\mathcal{V}}$ as the input into diffusion models fall into the space of $|z| \times T/f_t \times H /f_s \times W/f_s$, where $z$ denotes the latent channel. Thanks to the superior performance of open-source Stable Diffusion (SD)~\cite{stablediffusion} models, it has been a common practice to leverage SD-VAE to process images and videos into compressed latent. Typically, as shown in (a) of \cref{fig:vae_net}, SD-VAE is built upon 2D convolution with $f_s=8$ and $f_t=1$.

\vspace{0.05in}
\noindent \textbf{\system charm on videos.} As adjacent frames share large similarities in pixel space, video signals inherently contain more redundancy than images. Therefore, we apply a more aggressive down-sampling strategy for video autoencoder. Specifically, we employ 3D convolutions and increase the number of autoencoder blocks, increasing $f_t$ and $f_s$ to 4 and 16, as shown in (b) of \cref{fig:vae_net}, respectively. Though the overall down-sampling factor increases by $16\times$, the compressed latent space suffices to achieve comparable pixel-space reconstruction quality, highlighting the potential for extremely shrinking video latent dimensions.

As for image-to-video generation, we argue that video latent can be further compressed as the content image contains rich visual details and substantial overlapping information.
In such a scenario, video latents can be further compressed to represent the motion information.
We can find similar spirits in the video codec~\cite{richardson2011h}, where carefully designed modules use reference frames as inputs to perform high-fidelity reconstruction of a given compressed code. Specifically, we propose to cast the \textit{\system} charm on video latent space, as shown in (c) of \cref{fig:vae_net}. Generally, we choose the middle frame $V_{T/2}$ as the content guidance, compress input videos by an extreme down-sampling factor with a 3D encoder, and reconstruct from the latent space with the aid of content frame features pyramids. Specifically, we infuse content features of sizes varying in $H/8 \times W/8$ and $H/4 \times W/4$ via cross-attention within the decoder, adopt an extreme compression ratio, \ie, $f_s=32$ and $f_t=4$. While \textit{Reducio}-VAE projects videos into $64\times$ more compressed latent space than one of SDXL-VAE~\cite{podell2023sdxl}, it significantly outperforms the latter in reconstruction performance.

Note that we train \system-VAE on videos of $256 \times256$ resolution. To encode and decode high-resolution videos without exceeding the memory limit, we split the input video into overlapping tiles along the spatial dimension, following the approach used in Movie Gen~\cite{polyak2024movie}. In practice, we use a tile size of 256 (8 in latent space) with an overlap of 64 pixels (2 latents). Overlapping latents and decoded video regions are blended using linear combinations.

% \vspace{-0.5in}
\subsection{\textbf{\system} Diffusion Transformer}
\label{sec:reducio_dit}
Thanks to the magical compression capability of \textit{Reducio}-VAE, we project input videos into compressed latent space with a down-sampling factor of 4096, hence significantly speeding up the training and inference of diffusion models. 
Specifically, we adopt the DiT-XL~\cite{peebles2023scalable} model and follow most of the design choices adopted by PixArt-$\alpha$~\cite{chen2023pixart}, \ie, AdaLN-single modules, cross attention layers with Flan-T5-XXL~\cite{raffel2020exploring} text conditions. 
% The training target follows the reverse diffusion process~\cite{ddpm}, which can be depicted as
% \begin{equation}\small
% \label{eq:diffusion_reversion}
% \begin{aligned}
%      \bm{p}_{\theta}(\bm{z}_{t-1}|\bm{z}_{t}) = \mathcal{N}(\bm{z}_{t-1}; \bm{\mu}_{\theta}(\bm{z}_{t}, t), \bm{\sigma}_{t}^{2}\mathbf{I}),
% \end{aligned}
% \end{equation}
% where $\bm{z}$ is latent, $t$ is time step, $\bm{\mu}_{\theta}(\bm{z}_{t}, t)$ and $\bm{\sigma}_{t}^2$ indicate the mean and variance of the sample at the current time step.
% By re-parameterizing $\bm{\mu}_{\theta}$ as a noise prediction network $\varepsilon_\theta$, the model can be trained using simple mean-squared error between the predicted noise $\bm{\varepsilon}_{\theta}(\bm{z}_{t}, t)$ and the ground truth sampled Gaussian noise $\bm{\varepsilon}$ formulated as 
% \begin{equation}\small
% \label{eq:diffusion_loss}
% \begin{aligned}
%      \mathcal{L}_{simple} = \mathbb{E}_{\bm{z}_{0},\bm{\varepsilon},t}(\left \| \bm{\varepsilon} -\bm{\varepsilon}_{\theta}(\bm{z}_{t}, t)  \right \|^{2}).
% \end{aligned}
% \end{equation}

To adapt the image diffusion model to video, we consider two options: 
(1) Directly transforming the 2D attention to full 3D attention without adding additional parameters~\cite{yang2024cogvideox,chen2025goku}.
(2) Adding temporal layers and transforming the model to perform 2D spatial attention plus 1D temporal attention~\cite{zheng2024open,zeng2024make,lin2024stiv}.
We adopt option (1) by default. Studies on the evaluation of these two options are shown in Sec.~\ref{sec:exp_ablation_dit}.
In addition, we introduce \textit{Reducio}-DiT (illustrated in Fig.~\ref{fig:dit_net}) with additional image-condition modules, as described below.
% to tailor for image-conditioned video generation, 

\vspace{0.05in}
\noindent \textbf{Content frame modules} consist of a semantic encoder built upon pretrained OpenCLIP ViT-H~\cite{cherti2023reproducible} and a content encoder initialized with the SD2.1-VAE~\cite{stablediffusion}. The former encoder projects content frames into a high-level semantic space while the latter one mainly focuses on extracting the spatial information. 
We concatenate the obtained features with text tokens output by T5 to form the image-text-joint condition, and then cross-attend them with the noisy video latent. 

% Concretely, the detailed operations can be described as:
% \begin{equation}\small
% \label{eq:condition}
% \begin{aligned}
%      &\bm{e}_{s}=\mathtt{OpenCLIP}(V_{T/2}), \\
%      &\bm{e}_{c}=\mathtt{SD\mbox{-}VAE}(V_{T/2}), \\
%      &\bm{e}_{img}=\mathtt{MLP}(\bm{e}_{s})\oplus\mathtt{MLP}(\bm{e}_{c}), \\
%      &\bm{e}_{p}=\mathtt{MLP}(\mathtt{T5}(prompt)), \\
%      &\bm{e}=[\bm{e}_{img},\bm{e}_{p}], \\
%      &\bm{z}_{i}=\mathtt{CAttn}(\bm{z}_{i}, \bm{e}),
% \end{aligned}
% \end{equation}
% where $V_{T/2}$ is the condition image, $\oplus$ refers to element-wise addition, $[\cdot]$ refers to concatenation, and $\mathtt{CAttn}(\cdot)$ refers to cross-attention.

\begin{table*}[ht]
\small
\centering  
\caption{Quantitative comparison between \system-VAE and open-source state-of-the-art VAE for generation.}
\vspace{-0.1in}
\ra{1.15}
\addtolength{\tabcolsep}{7pt}
 \begin{tabular}{lccccccc}
 \toprule
 \multirow{2}{*}{\textbf{Model}} & {\textbf{Downsample}}  &\multirow{2}{*}{$|\bm{z}|$}  & \multicolumn{4}{c}{ \textbf{Pexels}}& \textbf{UCF-101} \\
 
 & {\textbf{Factor}}& & {\textbf{PSNR$\uparrow$}} & {\textbf{SSIM$\uparrow$}} &{\textbf{LPIPS$\downarrow$}}& {\textbf{rFVD$\downarrow$}} & \textbf{rFVD$\downarrow$}\\
 \cmidrule{1-8}
 SD2.1-VAE~\cite{stablediffusion} &$1 \times 8 \times 8$& 4 & 29.23 & 0.82 & 0.09 &25.96& \textbf{21.00} \\
 SDXL-VAE~\cite{podell2023sdxl} &$1 \times 8 \times 8$& 4 & 30.54	& 0.85 & 0.08 & 19.87 &23.68 \\
 OmniTokenizer~\cite{wang2024omnitokenizer} &$4 \times 8 \times 8$& 8 & 27.11 & 0.89 & 0.07 & 23.88 & 30.52\\
 OpenSora-1.2~\cite{zheng2024open} &$4 \times 8 \times 8$ & 16 &30.72 & 0.85 &	0.11 &60.88 & 67.52 \\
 \midrule
 \multirow{2}{*}{Cosmos-VAE~\cite{cosmos_vae}} & $8\times 8 \times 8$ & 16 & 30.84 & 0.74 & 0.12 & 29.44& 22.06 \\
 & $8\times 16 \times 16$ & 16 & 28.14 & 0.65 & 0.18 & 77.87 & 119.37 \\
 \textbf{Reducio-VAE} & $4 \times 32 \times 32$ & 16 & \textbf{35.88} & \textbf{0.94} & \textbf{0.05} & \textbf{17.88}& 65.17 \\
    \bottomrule
 \end{tabular}
 \label{tab:sota_vae}
\vspace{-0.1in}
\end{table*}

\begin{figure}[t]
\centering
% \vspace{-0.1in}
\includegraphics[width=\linewidth]{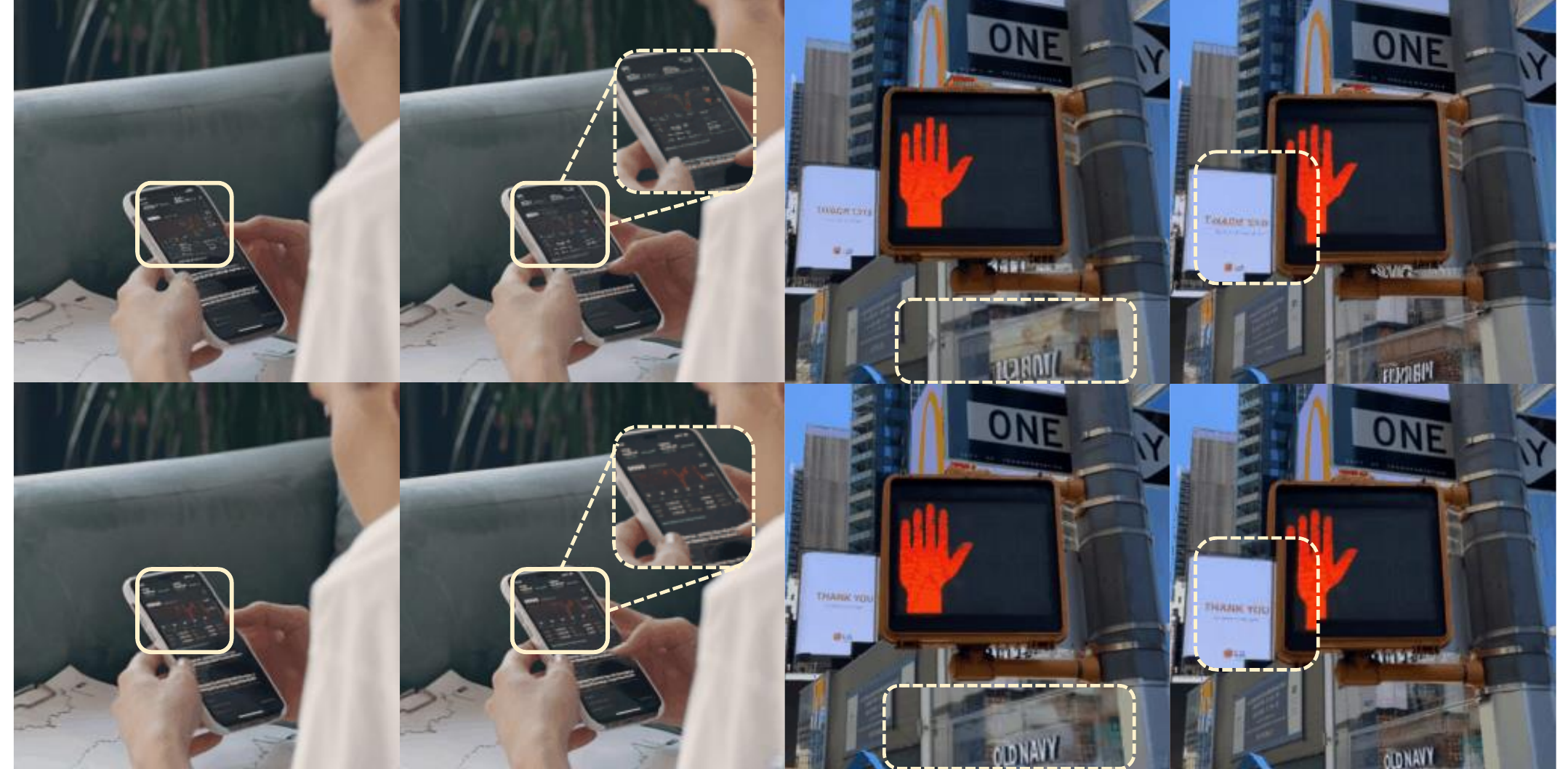}
\vspace{-0.15in}
\caption{Comparison between the $1^{st}$ and $16^{th}$ frames of videos reconstructed by SDXL-VAE (top) and \system-VAE (bottom). }
\label{fig:vae_vis_comp}
\vspace{-0.2in}
\end{figure}
\vspace{0.05in}
\noindent \textbf{Scaling up to high-resolution videos} consumes tremendous computational resources for vanilla video diffusion models. Nevertheless, \system-DiT relieves this limit to a great extent. To support high-resolution (\eg, $1024^2$) video generation, we employ a progressive training strategy: 1) in the first stage of training, the model learns to align the video latent space with text-image prior by taking in a vast number of $256^2$ videos as input; 2) in the second stage, we fine-tune the model on videos with higher resolution, \ie, $512 \times 512$. In consequence, the content encoder augments $4\times$ more tokens with the noisy latent; 3) in the third stage, we conduct further fine-tuning on videos with spatial resolution around $1024$ with multi-aspect augmentation. 

To better accommodate the model to videos with varying aspect ratios, we inject the aspect ratio and size of input videos as embeddings~\cite{chen2023pixart} into the DiT model. We then add the dynamic size embedding with timestep embedding before being fed into the MLP. More importantly, for models in the first and the second stages, the content encoder cross-attends $H/16 \times W/16$ tokens with the noisy latent. As computational costs grow rapidly when resolution increases, we introduce a strategy to ensure the efficiency of \system-DiT on high-resolution videos. Specifically, we aim to shrink the content embedding by $2\times$ in the spatial dimension in each cross-attention layer. Inspired by Deepstack~\cite{meng2024deepstack}, we divide input tokens into 4 groups of grids and iteratively concatenate each group of $H/32 \times W/32$ tokens with the other condition tokens. 

% We display the detailed framework in the supplementary.

\section{Experiments}
\label{sec:exp}

\subsection{Training and Evaluation Details}
Our \system-VAE is trained on a 400K video dataset collected from Pexels\footnote{\url{https://www.pexels.com/}}.
Pexels contains a large number of free and high-quality stock videos, where each video is accompanied by a short text description. We train \system-VAE from scratch on $256\times 256$ videos with 16 FPS. 
To perform the VAE comparison, PSNR, SSIM~\cite{wang2004image}, LPIPS~\cite{zhang2018perceptual} and reconstruction FVD (rFVD)~\cite{unterthiner2018towards} are employed as the evaluation metrics.

% where the former two assess the video quality, LPIPS assesses the perceptual similarities, and rFVD assesses the temporal coherence.

\system-DiT is trained on the above Pexels dataset and an internal dataset with 5M videos containing high-resolution text-video pairs. As described in Sec.~\ref{sec:reducio_dit}, we adopt a multi-stage training strategy to train the model from low resolution to high resolution. We first train \system-DiT-256 using a batch size of 512 on 4 Nvidia A100 80G GPUs for around 900 A100 hours and initialize model weights with PixArt-$\alpha$-256~\cite{chen2023pixart}. Then we fine-tune on $512^2$ videos shortly for 300 A100 hours to obtain \system-DiT-512. In the third stage, we randomly sample video batches from 40 aspect ratio buckets, which is the same as the setting in PixArt-$\alpha$. We leverage 8 AMD MI300 GPUs to support a training batch size of 768 and fine-tune \system-DiT-1024 for 1000 GPU hours.

\begin{table}
\small
\centering  
\caption{Quantitative comparison between \system-DiT and state-of-the-art image-to-video LDMs. }
\vspace{-0.1in}
\addtolength{\tabcolsep}{8pt}
\ra{1.2}
\begin{tabular}{lcc}
 \toprule
 \multirow{2}{*}{\textbf{Model}} &  \multicolumn{2}{c}{\textbf{FVD$\downarrow$}}\\ & {\textbf{UCF-101}} &  {\textbf{MSR-VTT}} \\
 \midrule
VideoComposer~\cite{wang2023videocomposer} & 576.8 & 377.3\\ 
 I2VGen-XL~\cite{zhang2023i2vgen} & 571.1& - \\
DynamiCrafter~\cite{xing2025dynamicrafter} & 429.2 & 234.7 \\
 \system-DiT & 318.5 & 291.9 \\
 \bottomrule
 \end{tabular}
\vspace{-0.2in}
\label{tab:i2v_cmp}
\end{table}

We adopt DPM-Solver++~\cite{lu2022dpm} as the sampling algorithm for efficient inference and set the sampling step to 20.
For evaluation in \cref{tab:dit_sota}, we use PixArt-$\alpha$-256 and PixArt-$\alpha$-1024 to generate the content frame for $256^2$ and $1024^2$ videos, respectively. We denote out-of-memory as OOM and calculate the speed of generating 16-frame video clips on a single A100 80G GPU, including both text-to-image and text-image to video cost. For \system-DiT,  We report the FVD and IS scores of \system-DiT-512 under the zero-shot setting on UCF-101~\cite{ucf101} and MSR-VTT~\cite{msrvtt}.
Since our model performs image-to-video generation and requires a condition image prior for the text-to-video pipeline, we follow MicroCinema~\cite{wang2024microcinema} to use SDXL~\cite{podell2023sdxl} to generate the condition image and LLaVA-1.5~\cite{liu2024improved} to generate captions for FVD evaluations on UCF-101 and MSR-VTT. Additionally, we evaluate \system-DiT-512 on the recent video generation benchmark VBench~\cite{huang2024vbench}.

\begin{table*}[t]
\small
\centering  
\caption{Quantitative comparison between \system-DiT and state-of-the-art text-to-video LDMs. }
\vspace{-0.1in}
\ra{1.2}
\addtolength{\tabcolsep}{2.5pt}
 \begin{tabular}{lccccccc}
 \toprule
\multirow{2}{*}{\textbf{Model}} & \multirow{2}{*}{\textbf{Params}} & \multirow{2}{*}{\textbf{VBench}} & \multicolumn{2}{c}{\textbf{FVD$\downarrow$}} &\multicolumn{2}{c}{\textbf{Throughput (\#videos/sec)}} & \textbf{GPU Hours} \\
& & & {\textbf{UCF-101}}   & {\textbf{MSR-VTT}} &   {\textbf{$256\times 256$}} & {\textbf{$1024\times 1024$}}& A100 Training\\
 \midrule
 Make-A-Video~\cite{singer2022make} & - & - &367.23 & - & -& - & - \\
 CogVideo~\cite{hong2022cogvideo}  & 7.8B& 67.01 &701.59 & 1294  & 0.0023 & \textbf{OOM} & $>$368K \\
 Show-1~\cite{zhang2024show} & 1.7B & 78.93 &394.46 & 538.0 & 0.0016 & \textbf{OOM} & $>$56K\\
 Lavie~\cite{wang2023lavie} & 1B & 77.08 & 526.30 & - & 0.1838 & 0.0039 & - \\
 MicroCinema~\cite{wang2024microcinema} & 2B & - &342.86 & 377.4 & 0.0234 & \textbf{OOM} & $>$6K\\
 Lumiere~\cite{bar2024lumiere} & $>$2B & - & 332.49 & - & - & - & - \\
 \system-DiT & 1.2B & 81.39 &  318.50 & 291.9 & 0.9824 & 0.0650 & 3.2K\\
\bottomrule
 \end{tabular}
\label{tab:dit_sota}
\vspace{-0.15in}
\end{table*}

\begin{figure*}[t]
    \centering
    \begin{minipage}[t]{.65\textwidth}
        \centering
     \includegraphics[width=0.95\textwidth]{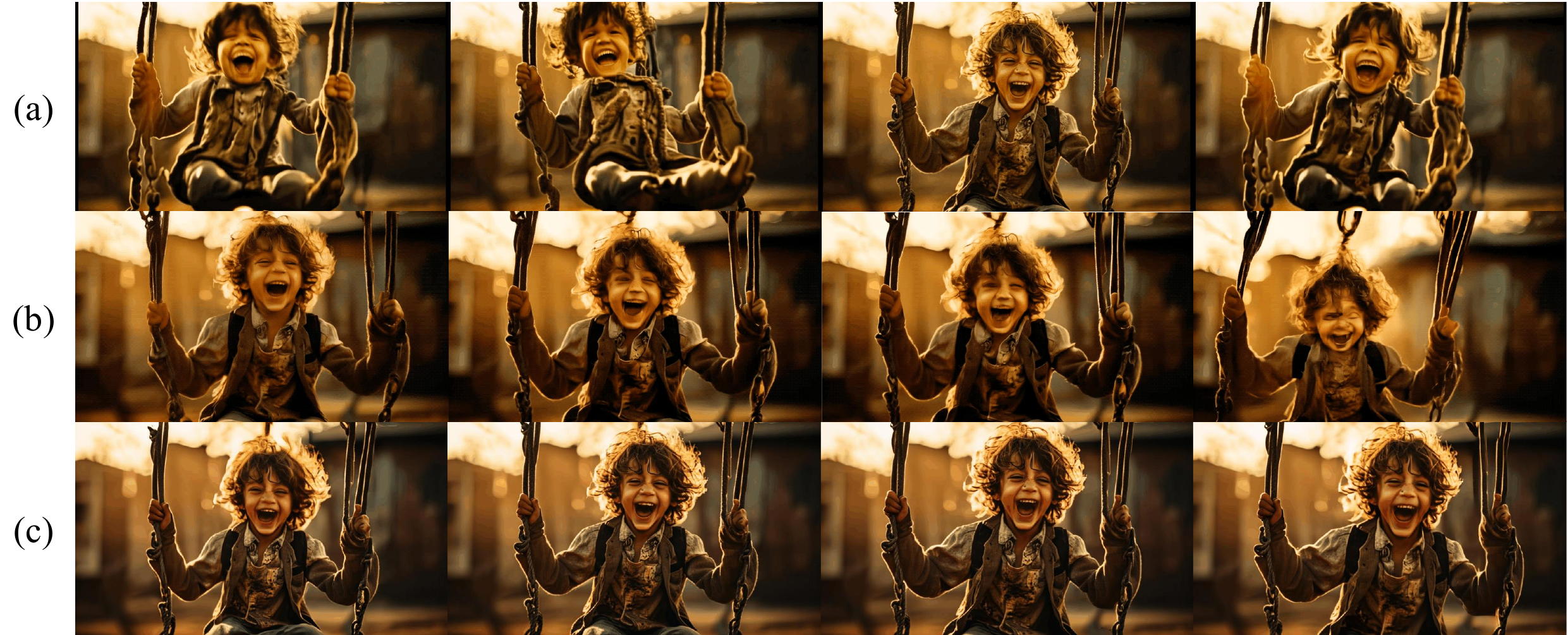}
     \textit{\small \blank{0.5cm} A child excitedly swings on a rusty swing set, laughter filling the air.}
     \vspace{0.03in}
    \end{minipage}%
    \begin{minipage}[t]{0.35\textwidth}
    \centering
    \includegraphics[width=0.95\textwidth]{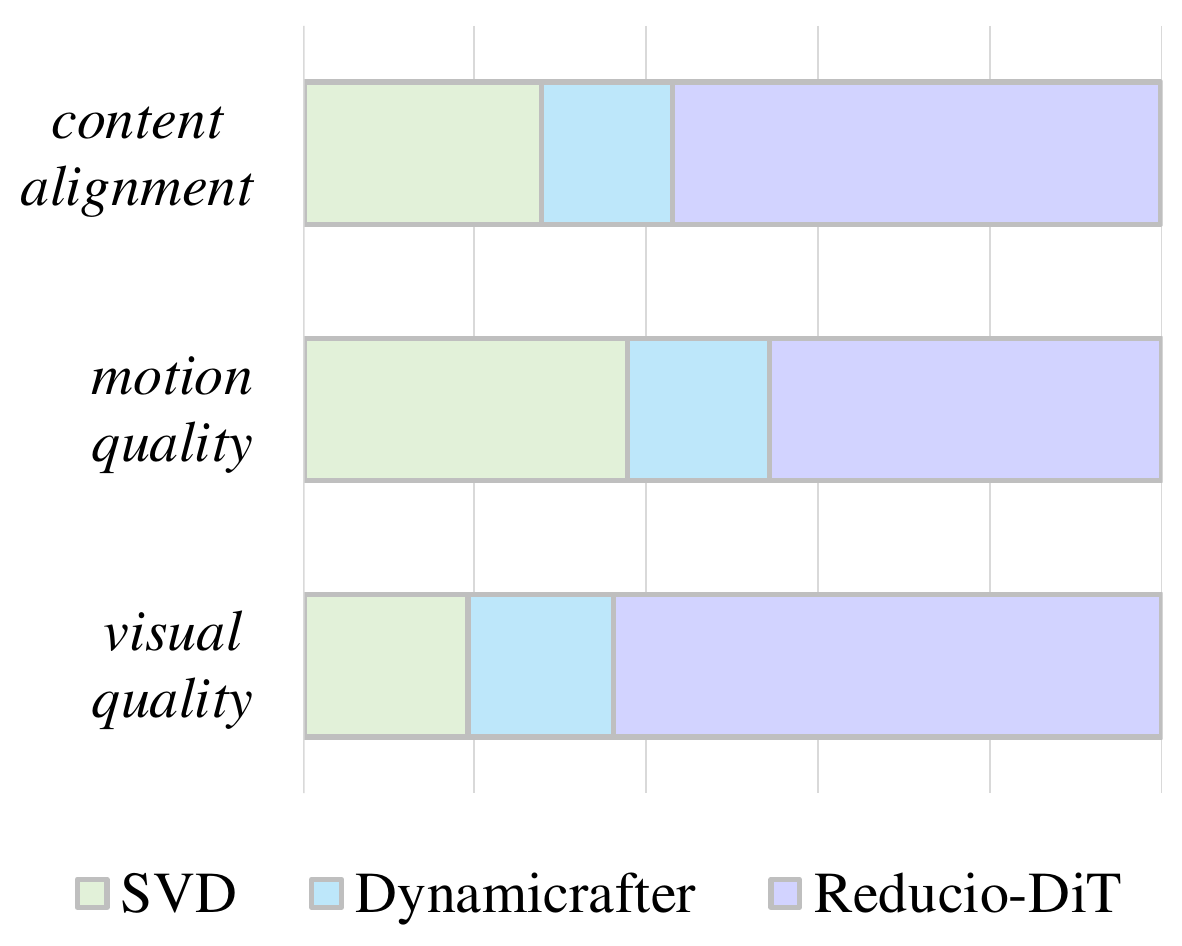}
    \end{minipage}\\
        \begin{minipage}[t]{0.65\textwidth}
        \centering
     \caption{Comparison between frames generated given an identical frame and prompt, by (a) DynamicCrafter~\cite{xing2025dynamicrafter}, (b) SVD-XT~\cite{blattmann2023stable} and (c) \system-DiT, respectively. We resize the output frames from $1344 \times 768$ to $1024\times576$ to match with baselines.}
     \label{fig:i2v_cmp}
     \vspace{-0.1in}
    \end{minipage}
    ~~
    \begin{minipage}[t]{0.3\textwidth}
        \centering
    % \hspace{0.01in}
     \caption{The user study of human preference rate for \system-DiT, DynamiCrafter~\cite{xing2025dynamicrafter} and SVD-XT~\cite{blattmann2023stable}.}
     \label{fig:user_study}
     \vspace{-0.1in}
    \end{minipage}
\vspace{-0.2in}
\end{figure*}

% GPU hours of other methods are estimated based on their iterations and GPU numbers. 

\subsection{Main Results}

\vspace{0.05in}
\noindent \textbf{\system-VAE preserves fine details} of video inputs. \cref{fig:vae_vis_comp} displays the first and the last frame of $256^2$ videos reconstructed by SDXL-VAE and \system-VAE, with the yellow bounding box highlighting the region with obvious differences in details. While SDXL-VAE causes corruptions and blurs, \system-VAE generally maintains the subtle texture within the middle frame.  
\cref{tab:sota_vae} also reflects the advantage of \system-VAE in quantitative reconstruction metrics, \eg, PSNR, SSIM, and LPIPS on a randomly sampled validation subset of 1K videos with the duration of 1s on Pexels, and test on both UCF-101 and Pexels for FVD. 

As shown in the table, \system-VAE strikes obvious advantages on overall reconstruction performance. Specifically, our model significantly outperforms state-of-the-art 2D VAE, \eg, SD2.1-VAE and SDXL-VAE, by more than 5 db in PSNR. Moreover, \system-VAE also performs better than VAEs designed for video in recent literature, \eg, OmniTokenizer and OpenSora-1.2. Our VAE also outperforms the concurrent work, Cosmos-VAE, which also adopts a further compressed video latent space, by 0.2 in SSIM and 5 db in PSNR with an $8\times$ higher down-sampling factor.

We observe that \system-VAE has worse rFVD than others on UCF-101. By careful inspection, we find the selected prior content image is blurring in many cases due to the low visual quality of UCF-101 and hence results in a lower reconstruction quality. In contrast, \system-VAE places first on rFVD on pexels, which consists of videos with much improved visual quality and much higher resolution. 

\vspace{0.05in}
\noindent \textbf{\system-DiT is a powerful image-to-video model.} As shown in \cref{tab:i2v_cmp}, \system-DiT achieves a 90-point lower FVD in the zero-shot setting on UCF-101 compared to the state-of-the-art image-to-video model, DynamiCrafter. Additionally, \system-DiT surpasses VideoComposer in FVD metrics on MSR-VTT but falls short of DynamiCrafter in this aspect. 
For qualitative evaluation, we conduct a user study using identical images generated by PixArt-$\alpha$-1024 as input. Participants evaluate 15 video groups based on three criteria: (1) visual quality, (2) motion quality, and (3) alignment of the condition image. Responses from 20 participants are presented in \cref{fig:user_study}. 

Notably, \system-DiT significantly outperforms in visual quality (64\%) and content alignment (57\%) by overcoming temporal flickering and preserving fine details from the condition image. Moreover, thanks to its strong temporal consistency, \system-DiT achieves a preference rate comparable to SVD-XT in motion quality, yet needs much fewer training hours and data. 
\cref{fig:i2v_cmp} displays the comparison between \system-DiT and the other two stat-of-the-art image-to-video models. While DynamicCrafter and SVD-XT fail to generate stable frames consistent with the given content frame, \system-DiT yields high visual quality with maintained facial details. Overall, \system-DiT demonstrates strong performance in image-to-video generation.

\begin{figure*}[t]
\centering
% \vspace{-0.1in}
\includegraphics[width=\linewidth]{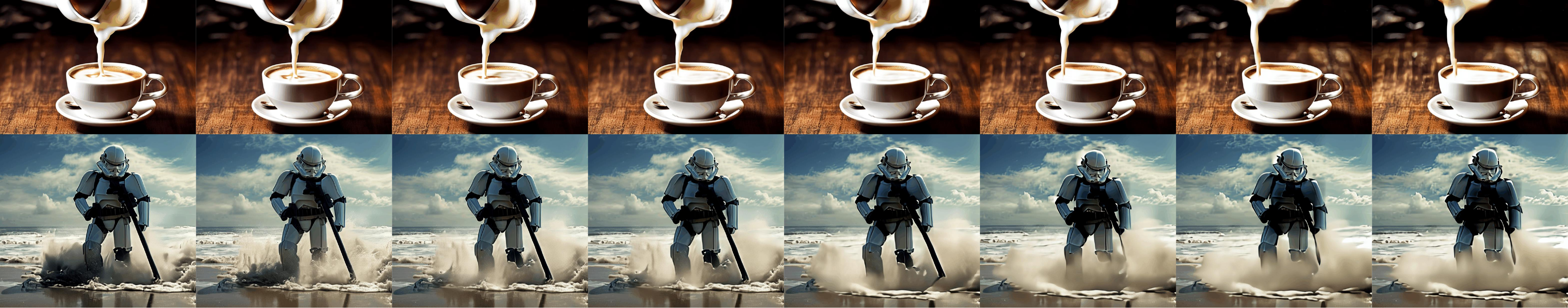}\\
\vspace{-0.03in}
\hspace{0.01in} \textbf{\small top:} \textit{\small 
pouring milk into a cup of coffee.
\blank{4cm}}  \textbf{\small bottom:} \textit{\small A storm trooper vacuuming the beach.}
\vspace{-0.05in}
\caption{Examples of a 2s video (top) and a 3s video (bottom) generated by \system-DiT in a recursive manner.}
\label{fig:long_video}
\vspace{-0.15in}
\end{figure*}

\vspace{0.05in}
\noindent \textbf{\system-DiT effectively balances efficiency and performance.} By integrating \system-DiT with a text-to-image LDM for condition image generation, the factorized text-to-video task can be performed efficiently. Consequently, our model achieves FVD scores of 318.5 on UCF-101 and 291.0 on MSR-VTT and the VBench score of 81.39, beating a range of state-of-the-art video LDMs, as recorded in \cref{tab:dit_sota}. More importantly, as most mainstream text-to-video models are built upon pre-trained text-to-image LDMs, we compare \system-DiT with baselines on training costs for video samples.
Thanks to the extremely compressed latent space, we achieve competitive performance with a training cost of merely 3.2K A100 hours.

Moreover, \system-DiT significantly boosts the throughput of video LDMs. The compressed latent space facilitates \system-DiT to output $256^2$ videos of 16 frames in almost 1 second, which realizes a $\mathbf{5.3\times}$ speed-up compared with Lavie. Moreover, 16-frame $1024^2$ video generation costs \system-DiT only for 15.4 seconds on average, leading to a striking speed-up of $\mathbf{16.6\times}$ over Lavie. We argue that the core philosophy of \system lies in the extremely compressed latent space, which accelerates both the training and inference speed of text-to-video generation.

Although \system-DiT is trained exclusively on 16-frame videos sampled at 16 fps, we present that it can be adapted for longer video generation in a zero-shot manner. This is achieved by recursively using the last frame of a generated clip as the input condition for the subsequent clip. \cref{fig:long_video} presents the visualization of frames sampled uniformly from 32-frame and 48-frame videos, demonstrating that \system-DiT maintains consistent motion while preserving high visual quality.

\subsection{Ablation Studies}
\subsubsection{Design of \textbf{\system}-VAE}
\begin{table}[t]
  \centering
  \small
  \caption{Ablation on the down-sampling factor of \system-VAE.} 
  \vspace{-0.05in}
     \addtolength{\tabcolsep}{13pt}
     \ra{1.2}
     % \small
   \resizebox{0.88\linewidth}{!}
   {\begin{tabular}{cccc}
    \toprule
   {\textbf{$f_t$}} & {\textbf{$f_s$}} & 
   {\textbf{PSNR$\uparrow$}} & {\textbf{SSIM$\uparrow$}} \\ 
    \cmidrule{1-4}
    16 & 16 & 34.14 & 0.93\\
    2 & 32 & 36.79 & 0.96 \\
    \rowcolor{lightblue} 4 & 32 & 35.88 & 0.94\\
    1 & 64 & 35.51 & 0.93\\ 
    \bottomrule
    \end{tabular}}
  \label{tab:vae_compress}
\vspace{-0.15in}
\end{table}

\vspace{0.05in}
\noindent \textbf{Spatial redundancy is the key for compression.} As \cref{tab:vae_compress} displays, when adopting an identical overall down-sampling factor $\mathcal{F} = f_t \cdot f_s^2$, \system-VAE achieves worse performance with a larger compression factor in the temporal dimension. The result suggests that when a reference frame is provided, videos contain heavier redundancy in the spatial dimension than in the temporal dimension. Similarly, when $f_s = 32$ is kept unchanged, increasing $f_t$ from 2 to 4 leads to a drop in PSNR by 0.91 db. To balance the trade-off between more compact latent space and better reconstruction performance, we set $f_t=4$ and $f_s=32$ in default.

\begin{figure}[t]
\centering
% \vspace{-0.1in}
\includegraphics[width=0.9\linewidth, trim=0 50 80 0, clip]{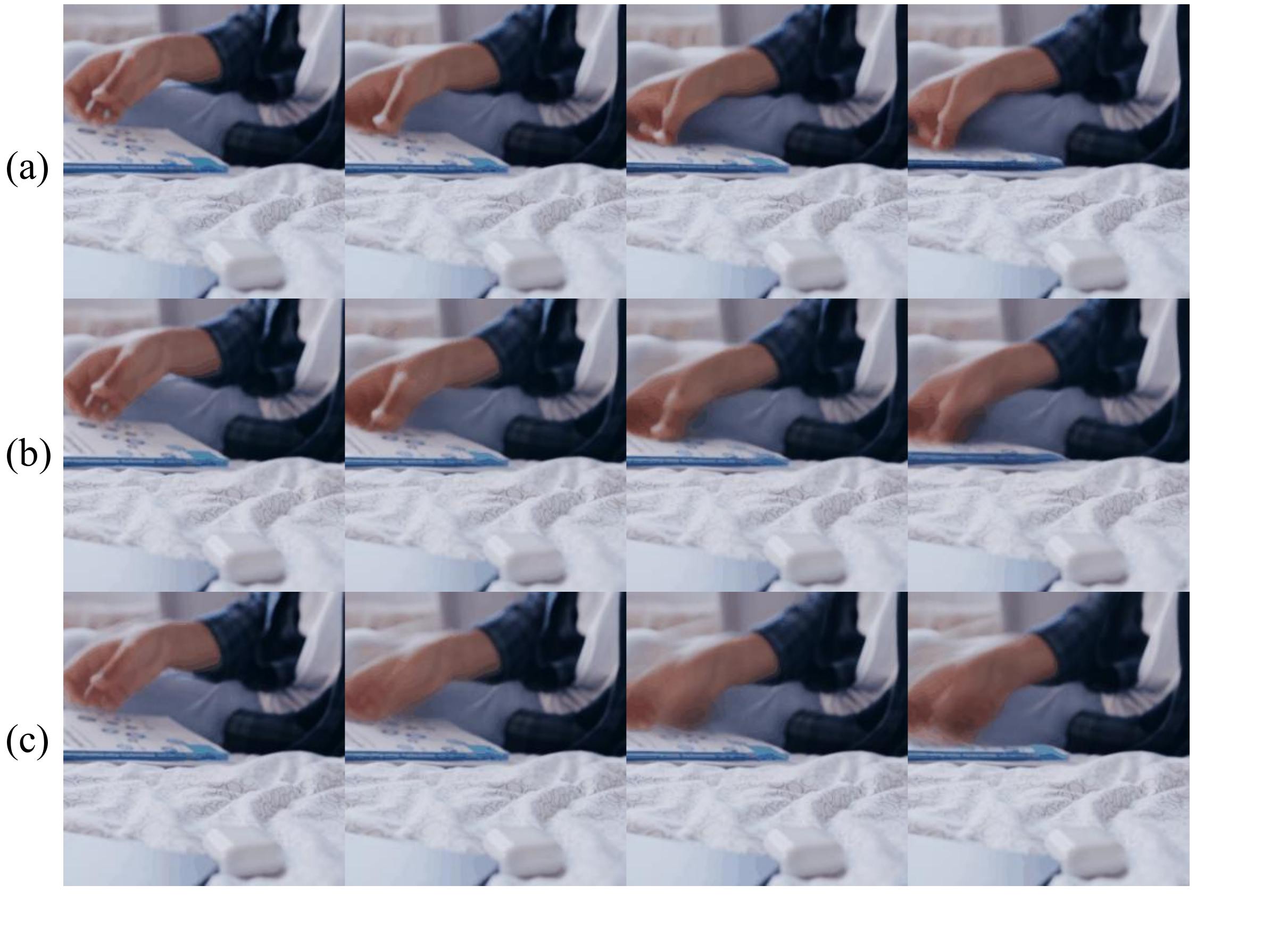}
\vspace{-0.1in}
\caption{Comparison between the $8^{\text{th}}$ to  $11^{\text{th}}$ frames of videos reconstructed by \system-VAE with different latent space channel dimensions, \ie, (a) $|\bm{z}|=16$, (b) $|\bm{z}|=8$, (c) $|\bm{z}|=4$.}
\label{fig:vae_z_comp}
\vspace{-0.15in}
\end{figure}

\vspace{0.05in}
\noindent \textbf{Scaling up latent channel is helpful} for improving the reconstruction performance, as results shown in \cref{tab:vae_channel}. When $\bm{z}$ grows from 4 to 16, the PSNR of reconstructed videos increases by 2.71 db. However, increasing the latent channel has a ceiling effect, as setting $\bm{z}$ to 32 hardly achieves further gains in both PSNR and SSIM. Therefore, we opt to use latent space with $|\bm{z}|=16$ for \system as the default setting.
A visual qualitative comparison is presented in \cref{fig:vae_z_comp}, where we compare the results when $|\bm{z}|=16$, $8$, and $4$.

\begin{table}[h]
  \small
\centering
\caption{Ablation on latent channel dimensions of  \system-VAE.}
\vspace{-0.05in}
\addtolength{\tabcolsep}{18pt}
\ra{1.2}
 % \small
\resizebox{0.88\linewidth}{!}
{\begin{tabular}{ccc}
\toprule
{$|\bm{z}|$} & {\textbf{PSNR$\uparrow$}} & {\textbf{SSIM$\uparrow$}} \\  
\cmidrule{1-3}
4 & 33.17 & 0.91 \\
8 & 34.15 & 0.93\\
\rowcolor{lightblue} 16 & 35.88 & 0.94\\
32 & 35.24 & 0.94\\
\bottomrule
\end{tabular}} 
\label{tab:vae_channel}
\vspace{-0.1in}
\end{table}

\vspace{0.05in}
\noindent \textbf{Fusion with cross-attention} achieves the most competitive performance in \cref{tab:vae_fuse}. As for add-based fusion, we feed the content frame features into a convolution and replicate the output on the temporal dimension. Then we add video features with the content feature, together with a temporal embedding. Concerning linear-based fusion, we view the content condition as a single-frame video feature and concatenate it with the video feature by the temporal dimension. A 3D convolution layer then refines the concatenated features. 
As observed in \cref{tab:vae_fuse}, the attention-based \system-VAE outperforms other baselines by 1.71 in PSNR and 0.01 in SSIM, while demanding higher computational costs. By default, we display results using the attention-based fusion. Note that all \system-VAE displayed in \cref{tab:vae_fuse} share the same 3D encoder and hence an identical latent space, where users are free to choose any of the decoders with different resource constraints.

\begin{table}[t]
  \small
    \centering
  \caption{Ablation on the fusion type in \system-VAE when infusing the content image.}
        \vspace{-0.05in}
        \addtolength{\tabcolsep}{6pt}
        \ra{1.2}
         \small
        \resizebox{0.88\linewidth}{!}
        {\begin{tabular}{cccc}
        \toprule
        {\textbf{Fusion}} & {\textbf{PSNR$\uparrow$}} & {\textbf{SSIM$\uparrow$}} & {\textbf{Params}} \\  
        \cmidrule{1-4}
        w/o. & 27.91 & 0.80 & 435M\\
        linear & 30.86 & 0.87 & 439M\\
        add & 34.17 & 0.93 & 450M \\
        \rowcolor{lightblue} attention & 35.88 & 0.94 & 437M\\
        \bottomrule
        \end{tabular}} 
\label{tab:vae_fuse}
\vspace{-0.15in}
\end{table}

\subsubsection{Design of \textbf{\system}-DiT}
\label{sec:exp_ablation_dit}
For ablation studies on DiT, we conduct training for 300 A100 hours for each experiment on Pexels and report FVD and IS scores on UCF-101. 

% In summary, compressing videos with \system-VAE is better than down-sampling latent with patch embedding. 

\vspace{0.05in}
\noindent \textbf{Using patch size of one} yields the best performance. As shown in \cref{tab:dit_pcz}, keeping a fixed spatial down-sampling factor of 32, we compare the performance of using a 2$\times$ higher temporal down-sampling factor in latent space and a temporal patch size of 1 ($f_t=4$, $p_t=1$), with temporal patch size of 2 when patchify the input latents for obtaining patch embedding ($f_t=2$, $p_t=2$).
%down-sampling the latent by 2 with the patch embedding in the temporal dimension. 
We observe that compressing the temporal dimension with VAE outperforms. Similarly, we ablate using VAE with 2$\times$ lower spatial down-sampling factor yet setting the patch size of spatial dimension to 2 ($f_s=16$, $p_s=2$). The experimental results identify that compressing the video with \system-VAE to an extreme extent and using a small patch size, \ie, 1, in both the temporal and spatial dimensions achieves better performance.

\begin{table}[h]
  \centering
  \small
\caption{Ablation on the temporal patch size $p_t$ and the spatial patch size $p_s$ in \system-DiT. } 
  \vspace{-0.05in}
     \addtolength{\tabcolsep}{4pt}
     \ra{1.2}
     % \small
   % \resizebox{0.85\linewidth}{!}
   {\begin{tabular}{cccccc}
    \toprule
   {\textbf{$f_t$}} & {\textbf{$f_s$}} & {\textbf{$p_t$}} & {\textbf{$p_s$}} &
   {\textbf{FVD$\downarrow$}}  &{\textbf{IS$\uparrow$}}\\ 
   \cmidrule{1-6}
    2 & 32 & 2 & 1 & 351.0 & 32.7\\
    4 & 16 & 1 & 2 & 534.1 & 30.4 \\
     \rowcolor{lightblue} 4 & 32 & 1 & 1 & 337.6 & 34.1\\
    \bottomrule
    \end{tabular}}
  \label{tab:dit_pcz}
\vspace{-0.1in}
\end{table}

% , which require heavy training to optimize the new parameters for obtaining good results. 

\begin{figure}[t]
\centering
% \vspace{-0.1in}
\includegraphics[width=\linewidth, trim=0 35 0 0, clip]{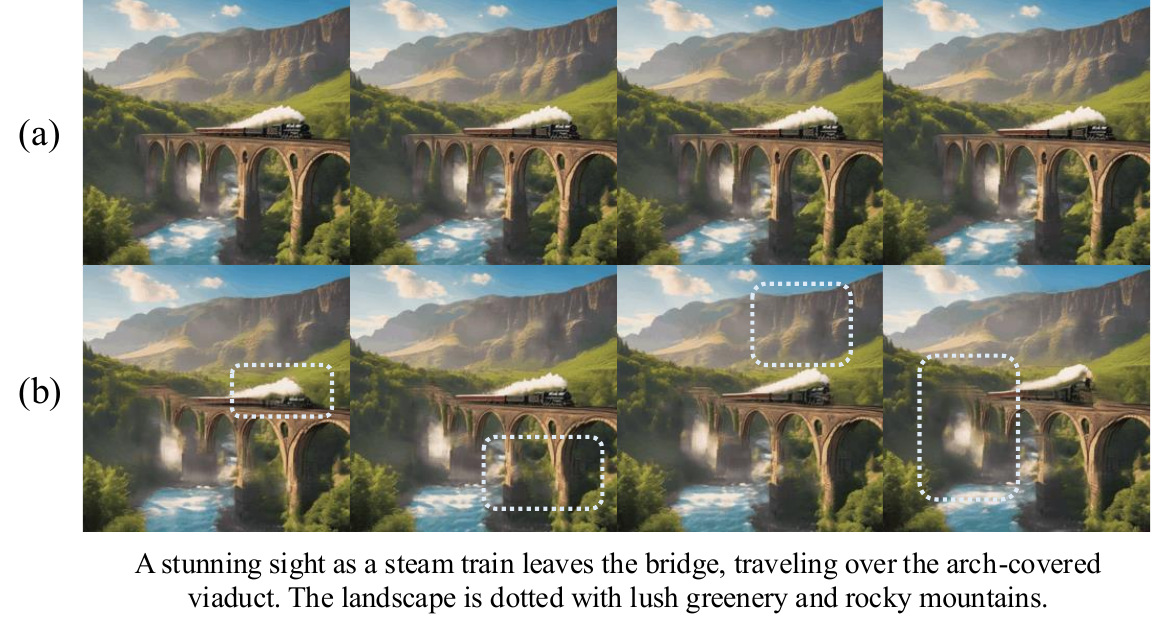}\\
\vspace{-0.1in}
\textit{\scriptsize A stunning sight as a steam train leaves the bridge, traveling over the arch-covered} \\
\vspace{-0.05in}
\textit{\scriptsize viaduct. The landscape is dotted with lush greenery and rocky mountains.}
\vspace{-0.04in}
\caption{Comparison on conditioning type in \system-DiT. (a) using semant.+content features. (b) using semantic features only.}
\label{fig:dit_content}
\vspace{-0.15in}
\end{figure}

\vspace{0.05in}
\noindent \textbf{Fusion with both semantic and content information} contributes to the best quality. As shown in \cref{tab:dit_cond}, using semantic features, \ie, OpenClip-based features alone may contribute to distorted visual detail, leading to a lower FVD. on the other hand, using content-based features helps to make video smooth and stable, achieving a higher FVD yet less versatile motion. 
A comparison example is illustrated in \cref{fig:dit_content}.
The collaboration of content and semantic features helps \system strike the best balance between content consistency and motion richness.

\begin{table}[h]
    \small
    \centering
    \caption{Ablation on the conditioning type in \system-DiT.}
        \vspace{-0.05in}
        \addtolength{\tabcolsep}{10pt}
        \ra{1.2}
         \small
        \resizebox{0.88\linewidth}{!}{\begin{tabular}{ccc}
        \toprule
        {\textbf{Condition}} & {\textbf{FVD$\downarrow$}} & {\textbf{IS$\uparrow$}} \\  
        \cmidrule{1-3}
        semantic & 472.1 & 34.5\\
        content & 363.7 & 31.1 \\
        \rowcolor{lightblue} semantic + content & 337.6 & 34.1\\
        \bottomrule
        \end{tabular}} 
        \label{tab:dit_cond}
\vspace{-0.15in}
\end{table}

\section{Conclusion}
Video generation has shown promising results with many potential applications while it is still trapped by the unaffordable computation cost.
In this paper, we explored how to effectively cut the overhead by reducing the latent size.
In particular, we found that video can be encoded into extremely compressed latents with the help of a content image prior, where the latent codes only need to represent the motion variables. 
With this observation, we designed the \system-VAE to compress videos to $4096\times$ smaller latents.
Using this powerful \system-VAE, we trained the \system-DiT that enables fast high-resolution video generation.
Our method is also compatible with other accelerating techniques, \eg, efficient attention, rectified flow, diffusion distillation, and \etc, allowing for further speedup, where we leave it for future exploration.

\vspace{0.025in}
\noindent\textbf{Limitations and future work.} While \system-DiT maintains strong consistency with the center frame, the number of training videos is still relatively limited for good instruction following performance. Also, 16-frame training samples somehow limit the magnitude of the motion in generated videos. As \system is much more compute-friendly with increased video frames and extended training data. Therefore, we believe in the potential of adapting our work for longer video generation on large-scale training and will explore it as future work.

\vspace{0.025in}
\noindent\textbf{Acknowledgement} This work was supported in part by the National Natural Science Foundation of China (Grant 62032006 and Grant 62472098).

% (16 frames of 16 fps, \ie, 1 second). In subsequence, 

{
    \small
    \bibliographystyle{ieeenat_fullname}

\begin{thebibliography}{85}
\providecommand{\natexlab}[1]{#1}
\providecommand{\url}[1]{\texttt{#1}}
\expandafter\ifx\csname urlstyle\endcsname\relax
  \providecommand{\doi}[1]{doi: #1}\else
  \providecommand{\doi}{doi: \begingroup \urlstyle{rm}\Url}\fi

\bibitem[An et~al.(2023)An, Zhang, Yang, Gupta, Huang, Luo, and Yin]{latentshift}
Jie An, Songyang Zhang, Harry Yang, Sonal Gupta, Jia-Bin Huang, Jiebo Luo, and Xi Yin.
\newblock Latent-shift: Latent diffusion with temporal shift for efficient text-to-video generation.
\newblock \emph{arXiv preprint arXiv:2304.08477}, 2023.

\bibitem[Bar-Tal et~al.(2024)Bar-Tal, Chefer, Tov, Herrmann, Paiss, Zada, Ephrat, Hur, Liu, Raj, et~al.]{bar2024lumiere}
Omer Bar-Tal, Hila Chefer, Omer Tov, Charles Herrmann, Roni Paiss, Shiran Zada, Ariel Ephrat, Junhwa Hur, Guanghui Liu, Amit Raj, et~al.
\newblock Lumiere: A space-time diffusion model for video generation.
\newblock In \emph{SIGGRAPH Asia}, 2024.

\bibitem[Blattmann et~al.(2023{\natexlab{a}})Blattmann, Dockhorn, Kulal, Mendelevitch, Kilian, Lorenz, Levi, English, Voleti, Letts, et~al.]{blattmann2023stable}
Andreas Blattmann, Tim Dockhorn, Sumith Kulal, Daniel Mendelevitch, Maciej Kilian, Dominik Lorenz, Yam Levi, Zion English, Vikram Voleti, Adam Letts, et~al.
\newblock Stable video diffusion: Scaling latent video diffusion models to large datasets.
\newblock \emph{arXiv preprint arXiv:2311.15127}, 2023{\natexlab{a}}.

\bibitem[Blattmann et~al.(2023{\natexlab{b}})Blattmann, Rombach, Ling, Dockhorn, Kim, Fidler, and Kreis]{blattmann2023align}
Andreas Blattmann, Robin Rombach, Huan Ling, Tim Dockhorn, Seung~Wook Kim, Sanja Fidler, and Karsten Kreis.
\newblock Align your latents: High-resolution video synthesis with latent diffusion models.
\newblock In \emph{CVPR}, 2023{\natexlab{b}}.

\bibitem[Chen et~al.(2023)Chen, Xia, He, Zhang, Cun, Yang, Xing, Liu, Chen, Wang, et~al.]{chen2023videocrafter1}
Haoxin Chen, Menghan Xia, Yingqing He, Yong Zhang, Xiaodong Cun, Shaoshu Yang, Jinbo Xing, Yaofang Liu, Qifeng Chen, Xintao Wang, et~al.
\newblock Videocrafter1: Open diffusion models for high-quality video generation.
\newblock \emph{arXiv preprint arXiv:2310.19512}, 2023.

\bibitem[Chen et~al.(2024)Chen, Yu, Ge, Yao, Xie, Wu, Wang, Kwok, Luo, Lu, et~al.]{chen2023pixart}
Junsong Chen, Jincheng Yu, Chongjian Ge, Lewei Yao, Enze Xie, Yue Wu, Zhongdao Wang, James Kwok, Ping Luo, Huchuan Lu, et~al.
\newblock Pixart-$\alpha$: Fast training of diffusion transformer for photorealistic text-to-image synthesis.
\newblock In \emph{ICLR}, 2024.

\bibitem[Chen et~al.(2025{\natexlab{a}})Chen, Cai, Chen, Xie, Yang, Tang, Li, Lu, and Han]{chen2024deep}
Junyu Chen, Han Cai, Junsong Chen, Enze Xie, Shang Yang, Haotian Tang, Muyang Li, Yao Lu, and Song Han.
\newblock Deep compression autoencoder for efficient high-resolution diffusion models.
\newblock In \emph{ICLR}, 2025{\natexlab{a}}.

\bibitem[Chen et~al.(2025{\natexlab{b}})Chen, Ge, Zhang, Zhang, Zhu, Yang, Hao, Wu, Lai, Hu, et~al.]{chen2025goku}
Shoufa Chen, Chongjian Ge, Yuqi Zhang, Yida Zhang, Fengda Zhu, Hao Yang, Hongxiang Hao, Hui Wu, Zhichao Lai, Yifei Hu, et~al.
\newblock Goku: Flow based video generative foundation models.
\newblock In \emph{CVPR}, 2025{\natexlab{b}}.

\bibitem[Cherti et~al.(2023)Cherti, Beaumont, Wightman, Wortsman, Ilharco, Gordon, Schuhmann, Schmidt, and Jitsev]{cherti2023reproducible}
Mehdi Cherti, Romain Beaumont, Ross Wightman, Mitchell Wortsman, Gabriel Ilharco, Cade Gordon, Christoph Schuhmann, Ludwig Schmidt, and Jenia Jitsev.
\newblock Reproducible scaling laws for contrastive language-image learning.
\newblock In \emph{CVPR}, 2023.

\bibitem[Esser et~al.(2024)Esser, Kulal, Blattmann, Entezari, M{\"u}ller, Saini, Levi, Lorenz, Sauer, Boesel, et~al.]{esser2024scaling}
Patrick Esser, Sumith Kulal, Andreas Blattmann, Rahim Entezari, Jonas M{\"u}ller, Harry Saini, Yam Levi, Dominik Lorenz, Axel Sauer, Frederic Boesel, et~al.
\newblock Scaling rectified flow transformers for high-resolution image synthesis.
\newblock In \emph{ICML}, 2024.

\bibitem[Fei et~al.(2024)Fei, Fan, Yu, Li, Zhang, and Huang]{fei2024dimba}
Zhengcong Fei, Mingyuan Fan, Changqian Yu, Debang Li, Youqiang Zhang, and Junshi Huang.
\newblock Dimba: Transformer-mamba diffusion models.
\newblock \emph{arXiv preprint arXiv:2406.01159}, 2024.

\bibitem[Feng et~al.(2024)Feng, Weng, Wang, Yuan, Bao, Luo, Chen, and Guo]{feng2024ccedit}
Ruoyu Feng, Wenming Weng, Yanhui Wang, Yuhui Yuan, Jianmin Bao, Chong Luo, Zhibo Chen, and Baining Guo.
\newblock Ccedit: Creative and controllable video editing via diffusion models.
\newblock In \emph{CVPR}, 2024.

\bibitem[Girdhar et~al.(2024)Girdhar, Singh, Brown, Duval, Azadi, Rambhatla, Shah, Yin, Parikh, and Misra]{girdhar2023emu}
Rohit Girdhar, Mannat Singh, Andrew Brown, Quentin Duval, Samaneh Azadi, Sai~Saketh Rambhatla, Akbar Shah, Xi Yin, Devi Parikh, and Ishan Misra.
\newblock Emu video: Factorizing text-to-video generation by explicit image conditioning.
\newblock In \emph{ECCV}, 2024.

\bibitem[Hang et~al.(2023)Hang, Gu, Li, Bao, Chen, Hu, Geng, and Guo]{hang2023efficient}
Tiankai Hang, Shuyang Gu, Chen Li, Jianmin Bao, Dong Chen, Han Hu, Xin Geng, and Baining Guo.
\newblock Efficient diffusion training via min-snr weighting strategy.
\newblock In \emph{ICCV}, 2023.

\bibitem[He et~al.(2024)He, Xue, Liu, Lin, Gao, Lin, Qiao, Ouyang, and Liu]{he2024venhancer}
Jingwen He, Tianfan Xue, Dongyang Liu, Xinqi Lin, Peng Gao, Dahua Lin, Yu Qiao, Wanli Ouyang, and Ziwei Liu.
\newblock Venhancer: Generative space-time enhancement for video generation.
\newblock \emph{arXiv preprint arXiv:2407.07667}, 2024.

\bibitem[Ho and Salimans(2022)]{ho2022classifier}
Jonathan Ho and Tim Salimans.
\newblock Classifier-free diffusion guidance.
\newblock \emph{arXiv preprint arXiv:2207.12598}, 2022.

\bibitem[Ho et~al.(2020)Ho, Jain, and Abbeel]{ddpm}
Jonathan Ho, Ajay Jain, and Pieter Abbeel.
\newblock Denoising diffusion probabilistic models.
\newblock In \emph{NeurIPS}, 2020.

\bibitem[Hong et~al.(2023)Hong, Ding, Zheng, Liu, and Tang]{hong2022cogvideo}
Wenyi Hong, Ming Ding, Wendi Zheng, Xinghan Liu, and Jie Tang.
\newblock Cogvideo: Large-scale pretraining for text-to-video generation via transformers.
\newblock \emph{ICLR}, 2023.

\bibitem[Hu et~al.(2025)Hu, Chen, and Luo]{hu2023lamd}
Yaosi Hu, Zhenzhong Chen, and Chong Luo.
\newblock Lamd: Latent motion diffusion for video generation.
\newblock \emph{IJCV}, 2025.

\bibitem[Huang et~al.(2024)Huang, He, Yu, Zhang, Si, Jiang, Zhang, Wu, Jin, Chanpaisit, et~al.]{huang2024vbench}
Ziqi Huang, Yinan He, Jiashuo Yu, Fan Zhang, Chenyang Si, Yuming Jiang, Yuanhan Zhang, Tianxing Wu, Qingyang Jin, Nattapol Chanpaisit, et~al.
\newblock Vbench: Comprehensive benchmark suite for video generative models.
\newblock In \emph{CVPR}, 2024.

\bibitem[Isola et~al.(2017)Isola, Zhu, Zhou, and Efros]{isola2017image}
Phillip Isola, Jun-Yan Zhu, Tinghui Zhou, and Alexei~A Efros.
\newblock Image-to-image translation with conditional adversarial networks.
\newblock In \emph{CVPR}, 2017.

\bibitem[Jin et~al.(2025)Jin, Sun, Li, Xu, Jiang, Zhuang, Huang, Song, Mu, and Lin]{jin2024pyramidal}
Yang Jin, Zhicheng Sun, Ningyuan Li, Kun Xu, Hao Jiang, Nan Zhuang, Quzhe Huang, Yang Song, Yadong Mu, and Zhouchen Lin.
\newblock Pyramidal flow matching for efficient video generative modeling.
\newblock In \emph{ICLR}, 2025.

\bibitem[Kingma and Welling(2014)]{vae}
Diederik~P Kingma and Max Welling.
\newblock Auto-encoding variational bayes.
\newblock In \emph{ICLR}, 2014.

\bibitem[Kuaishou(2024)]{kling}
Kuaishou.
\newblock Kling.
\newblock \url{https://kling.kuaishou.com/en}, 2024.

\bibitem[Labs(2014)]{flux2024}
Black~Forest Labs.
\newblock Flux.1.
\newblock \url{https://github.com/black-forest-labs/flux}, 2014.

\bibitem[Lin et~al.(2024)Lin, Liu, Chen, Lu, Hu, Fu, Allardice, Lai, Song, Zhang, et~al.]{lin2024stiv}
Zongyu Lin, Wei Liu, Chen Chen, Jiasen Lu, Wenze Hu, Tsu-Jui Fu, Jesse Allardice, Zhengfeng Lai, Liangchen Song, Bowen Zhang, et~al.
\newblock Stiv: Scalable text and image conditioned video generation.
\newblock \emph{arXiv preprint arXiv:2412.07730}, 2024.

\bibitem[Liu et~al.(2024)Liu, Li, Li, and Lee]{liu2024improved}
Haotian Liu, Chunyuan Li, Yuheng Li, and Yong~Jae Lee.
\newblock Improved baselines with visual instruction tuning.
\newblock In \emph{CVPR}, 2024.

\bibitem[Liu et~al.(2023)Liu, Zhang, Ma, Peng, et~al.]{liu2023instaflow}
Xingchao Liu, Xiwen Zhang, Jianzhu Ma, Jian Peng, et~al.
\newblock Instaflow: One step is enough for high-quality diffusion-based text-to-image generation.
\newblock In \emph{ICLR}, 2023.

\bibitem[Lu et~al.(2022)Lu, Zhou, Bao, Chen, Li, and Zhu]{lu2022dpm}
Cheng Lu, Yuhao Zhou, Fan Bao, Jianfei Chen, Chongxuan Li, and Jun Zhu.
\newblock Dpm-solver++: Fast solver for guided sampling of diffusion probabilistic models.
\newblock \emph{arXiv preprint arXiv:2211.01095}, 2022.

\bibitem[Luhman and Luhman(2021)]{luhman2021knowledge}
Eric Luhman and Troy Luhman.
\newblock Knowledge distillation in iterative generative models for improved sampling speed.
\newblock \emph{arXiv preprint arXiv:2101.02388}, 2021.

\bibitem[Ma et~al.(2025)Ma, Sun, Ma, Tang, Ma, Wang, Li, Dai, Shi, Ju, et~al.]{ma2025token}
Xu Ma, Peize Sun, Haoyu Ma, Hao Tang, Chih-Yao Ma, Jialiang Wang, Kunpeng Li, Xiaoliang Dai, Yujun Shi, Xuan Ju, et~al.
\newblock Token-shuffle: Towards high-resolution image generation with autoregressive models.
\newblock \emph{arXiv preprint arXiv:2504.17789}, 2025.

\bibitem[Meng et~al.(2024)Meng, Yang, Tian, Dai, Wu, Gao, and Jiang]{meng2024deepstack}
Lingchen Meng, Jianwei Yang, Rui Tian, Xiyang Dai, Zuxuan Wu, Jianfeng Gao, and Yu-Gang Jiang.
\newblock Deepstack: Deeply stacking visual tokens is surprisingly simple and effective for lmms.
\newblock In \emph{NeurIPS}, 2024.

\bibitem[Nichol et~al.(2022)Nichol, Dhariwal, Ramesh, Shyam, Mishkin, McGrew, Sutskever, and Chen]{glide}
Alex Nichol, Prafulla Dhariwal, Aditya Ramesh, Pranav Shyam, Pamela Mishkin, Bob McGrew, Ilya Sutskever, and Mark Chen.
\newblock Glide: Towards photorealistic image generation and editing with text-guided diffusion models.
\newblock In \emph{ICML}, 2022.

\bibitem[Nichol and Dhariwal(2021)]{improvedddpm}
Alexander~Quinn Nichol and Prafulla Dhariwal.
\newblock Improved denoising diffusion probabilistic models.
\newblock In \emph{ICML}, 2021.

\bibitem[Nvidia(2024)]{cosmos_vae}
Nvidia.
\newblock Cosmos tokenizer: A suite of image and video neural tokenizers.
\newblock \url{https://github.com/NVIDIA/Cosmos-Tokenizer}, 2024.

\bibitem[OpenAI(2024)]{sora}
OpenAI.
\newblock Sora.
\newblock \url{https://openai.com/index/sora/}, 2024.

\bibitem[Peebles and Xie(2023)]{peebles2023scalable}
William Peebles and Saining Xie.
\newblock Scalable diffusion models with transformers.
\newblock In \emph{ICCV}, 2023.

\bibitem[Pernias et~al.(2024)Pernias, Rampas, Richter, Pal, and Aubreville]{perniaswurstchen}
Pablo Pernias, Dominic Rampas, Mats~Leon Richter, Christopher Pal, and Marc Aubreville.
\newblock W{\"u}rstchen: An efficient architecture for large-scale text-to-image diffusion models.
\newblock In \emph{ICLR}, 2024.

\bibitem[Podell et~al.(2024)Podell, English, Lacey, Blattmann, Dockhorn, M{\"u}ller, Penna, and Rombach]{podell2023sdxl}
Dustin Podell, Zion English, Kyle Lacey, Andreas Blattmann, Tim Dockhorn, Jonas M{\"u}ller, Joe Penna, and Robin Rombach.
\newblock Sdxl: Improving latent diffusion models for high-resolution image synthesis.
\newblock In \emph{ICLR}, 2024.

\bibitem[Polyak et~al.(2024)Polyak, Zohar, Brown, Tjandra, Sinha, Lee, Vyas, Shi, Ma, Chuang, et~al.]{polyak2024movie}
Adam Polyak, Amit Zohar, Andrew Brown, Andros Tjandra, Animesh Sinha, Ann Lee, Apoorv Vyas, Bowen Shi, Chih-Yao Ma, Ching-Yao Chuang, et~al.
\newblock Movie gen: A cast of media foundation models.
\newblock \emph{arXiv preprint arXiv:2410.13720}, 2024.

\bibitem[Qing et~al.(2024)Qing, Zhang, Wang, Wang, Wei, Zhang, Gao, and Sang]{qing2024hierarchical}
Zhiwu Qing, Shiwei Zhang, Jiayu Wang, Xiang Wang, Yujie Wei, Yingya Zhang, Changxin Gao, and Nong Sang.
\newblock Hierarchical spatio-temporal decoupling for text-to-video generation.
\newblock In \emph{CVPR}, 2024.

\bibitem[Raffel et~al.(2020)Raffel, Shazeer, Roberts, Lee, Narang, Matena, Zhou, Li, and Liu]{raffel2020exploring}
Colin Raffel, Noam Shazeer, Adam Roberts, Katherine Lee, Sharan Narang, Michael Matena, Yanqi Zhou, Wei Li, and Peter~J Liu.
\newblock Exploring the limits of transfer learning with a unified text-to-text transformer.
\newblock \emph{JMLR}, 2020.

\bibitem[Rezende et~al.(2014)Rezende, Mohamed, and Wierstra]{rezende2014stochastic}
Danilo~Jimenez Rezende, Shakir Mohamed, and Daan Wierstra.
\newblock Stochastic backpropagation and approximate inference in deep generative models.
\newblock In \emph{ICML}, 2014.

\bibitem[Richardson(2011)]{richardson2011h}
Iain~E Richardson.
\newblock \emph{The H. 264 advanced video compression standard}.
\newblock John Wiley \& Sons, 2011.

\bibitem[Rombach et~al.(2022)Rombach, Blattmann, Lorenz, Esser, and Ommer]{stablediffusion}
Robin Rombach, Andreas Blattmann, Dominik Lorenz, Patrick Esser, and Bj{\"o}rn Ommer.
\newblock High-resolution image synthesis with latent diffusion models.
\newblock In \emph{CVPR}, 2022.

\bibitem[Runway(2024)]{gen3}
Runway.
\newblock Gen-3.
\newblock \url{https://runwayml.com/research/introducing-gen-3-alpha}, 2024.

\bibitem[Singer et~al.(2023)Singer, Polyak, Hayes, Yin, An, Zhang, Hu, Yang, Ashual, Gafni, et~al.]{singer2022make}
Uriel Singer, Adam Polyak, Thomas Hayes, Xi Yin, Jie An, Songyang Zhang, Qiyuan Hu, Harry Yang, Oron Ashual, Oran Gafni, et~al.
\newblock Make-a-video: Text-to-video generation without text-video data.
\newblock \emph{ICLR}, 2023.

\bibitem[Song et~al.(2021)Song, Sohl-Dickstein, Kingma, Kumar, Ermon, and Poole]{song2020score}
Yang Song, Jascha Sohl-Dickstein, Diederik~P Kingma, Abhishek Kumar, Stefano Ermon, and Ben Poole.
\newblock Score-based generative modeling through stochastic differential equations.
\newblock In \emph{ICLR}, 2021.

\bibitem[Song et~al.(2023)Song, Dhariwal, Chen, and Sutskever]{song2023consistency}
Yang Song, Prafulla Dhariwal, Mark Chen, and Ilya Sutskever.
\newblock Consistency models.
\newblock In \emph{ICML}, 2023.

\bibitem[Soomro et~al.(2012)Soomro, Zamir, and Shah]{ucf101}
Khurram Soomro, Amir~Roshan Zamir, and Mubarak Shah.
\newblock Ucf101: A dataset of 101 human actions classes from videos in the wild.
\newblock \emph{arXiv preprint arXiv:1212.0402}, 2012.

\bibitem[Tang et~al.(2025)Tang, Wu, Yang, Xie, Chen, Chen, Zhang, Cai, Lu, and Han]{tang2024hart}
Haotian Tang, Yecheng Wu, Shang Yang, Enze Xie, Junsong Chen, Junyu Chen, Zhuoyang Zhang, Han Cai, Yao Lu, and Song Han.
\newblock Hart: Efficient visual generation with hybrid autoregressive transformer.
\newblock In \emph{ICLR}, 2025.

\bibitem[Teng et~al.(2024)Teng, Wu, Shi, Ning, Dai, Wang, Li, and Liu]{teng2024dim}
Yao Teng, Yue Wu, Han Shi, Xuefei Ning, Guohao Dai, Yu Wang, Zhenguo Li, and Xihui Liu.
\newblock Dim: Diffusion mamba for efficient high-resolution image synthesis.
\newblock \emph{arXiv preprint arXiv:2405.14224}, 2024.

\bibitem[Tu et~al.(2023)Tu, Dai, Wu, Cheng, Hu, and Jiang]{tu2023implicit}
Shuyuan Tu, Qi Dai, Zuxuan Wu, Zhi-Qi Cheng, Han Hu, and Yu-Gang Jiang.
\newblock Implicit temporal modeling with learnable alignment for video recognition.
\newblock In \emph{ICCV}, 2023.

\bibitem[Tu et~al.(2024{\natexlab{a}})Tu, Dai, Cheng, Hu, Han, Wu, and Jiang]{tu2024motioneditor}
Shuyuan Tu, Qi Dai, Zhi-Qi Cheng, Han Hu, Xintong Han, Zuxuan Wu, and Yu-Gang Jiang.
\newblock Motioneditor: Editing video motion via content-aware diffusion.
\newblock In \emph{CVPR}, 2024{\natexlab{a}}.

\bibitem[Tu et~al.(2024{\natexlab{b}})Tu, Dai, Zhang, Xie, Cheng, Luo, Han, Wu, and Jiang]{tu2024motionfollower}
Shuyuan Tu, Qi Dai, Zihao Zhang, Sicheng Xie, Zhi-Qi Cheng, Chong Luo, Xintong Han, Zuxuan Wu, and Yu-Gang Jiang.
\newblock Motionfollower: Editing video motion via lightweight score-guided diffusion.
\newblock \emph{arXiv preprint arXiv:2405.20325}, 2024{\natexlab{b}}.

\bibitem[Tu et~al.(2025)Tu, Xing, Han, Cheng, Dai, Luo, and Wu]{tu2025stableanimator}
Shuyuan Tu, Zhen Xing, Xintong Han, Zhi-Qi Cheng, Qi Dai, Chong Luo, and Zuxuan Wu.
\newblock Stableanimator: High-quality identity-preserving human image animation.
\newblock In \emph{CVPR}, 2025.

\bibitem[Unterthiner et~al.(2018)Unterthiner, Van~Steenkiste, Kurach, Marinier, Michalski, and Gelly]{unterthiner2018towards}
Thomas Unterthiner, Sjoerd Van~Steenkiste, Karol Kurach, Raphael Marinier, Marcin Michalski, and Sylvain Gelly.
\newblock Towards accurate generative models of video: A new metric \& challenges.
\newblock \emph{arXiv preprint arXiv:1812.01717}, 2018.

\bibitem[Wan et~al.(2025)Wan, Wang, Ai, Wen, Mao, Xie, Chen, Yu, Zhao, Yang, et~al.]{wan2025wan}
Team Wan, Ang Wang, Baole Ai, Bin Wen, Chaojie Mao, Chen-Wei Xie, Di Chen, Feiwu Yu, Haiming Zhao, Jianxiao Yang, et~al.
\newblock Wan: Open and advanced large-scale video generative models.
\newblock \emph{arXiv preprint arXiv:2503.20314}, 2025.

\bibitem[Wang et~al.(2024{\natexlab{a}})Wang, Jiang, Yuan, Peng, Wu, and Jiang]{wang2024omnitokenizer}
Junke Wang, Yi Jiang, Zehuan Yuan, Binyue Peng, Zuxuan Wu, and Yu-Gang Jiang.
\newblock Omnitokenizer: A joint image-video tokenizer for visual generation.
\newblock In \emph{NeurIPS}, 2024{\natexlab{a}}.

\bibitem[Wang et~al.(2025{\natexlab{a}})Wang, Tian, Wang, Zhang, Huang, Wu, and Jiang]{wang2025simplear}
Junke Wang, Zhi Tian, Xun Wang, Xinyu Zhang, Weilin Huang, Zuxuan Wu, and Yu-Gang Jiang.
\newblock Simplear: Pushing the frontier of autoregressive visual generation through pretraining, sft, and rl.
\newblock \emph{arXiv preprint arXiv:2504.11455}, 2025{\natexlab{a}}.

\bibitem[Wang et~al.(2025{\natexlab{b}})Wang, Yang, Tuo, He, Zhu, Fu, and Liu]{videofactory}
Wenjing Wang, Huan Yang, Zixi Tuo, Huiguo He, Junchen Zhu, Jianlong Fu, and Jiaying Liu.
\newblock Swap attention in spatiotemporal diffusions for text-to-video generation.
\newblock \emph{IJCV}, 2025{\natexlab{b}}.

\bibitem[Wang et~al.(2023)Wang, Yuan, Zhang, Chen, Wang, Zhang, Shen, Zhao, and Zhou]{wang2023videocomposer}
Xiang Wang, Hangjie Yuan, Shiwei Zhang, Dayou Chen, Jiuniu Wang, Yingya Zhang, Yujun Shen, Deli Zhao, and Jingren Zhou.
\newblock Videocomposer: Compositional video synthesis with motion controllability.
\newblock In \emph{NeurIPS}, 2023.

\bibitem[Wang et~al.(2024{\natexlab{b}})Wang, Bao, Weng, Feng, Yin, Yang, Zhang, Dai, Zhao, Wang, et~al.]{wang2024microcinema}
Yanhui Wang, Jianmin Bao, Wenming Weng, Ruoyu Feng, Dacheng Yin, Tao Yang, Jingxu Zhang, Qi Dai, Zhiyuan Zhao, Chunyu Wang, et~al.
\newblock Microcinema: A divide-and-conquer approach for text-to-video generation.
\newblock In \emph{CVPR}, 2024{\natexlab{b}}.

\bibitem[Wang et~al.(2024{\natexlab{c}})Wang, Chen, Ma, Zhou, Huang, Wang, Yang, He, Yu, Yang, et~al.]{wang2023lavie}
Yaohui Wang, Xinyuan Chen, Xin Ma, Shangchen Zhou, Ziqi Huang, Yi Wang, Ceyuan Yang, Yinan He, Jiashuo Yu, Peiqing Yang, et~al.
\newblock Lavie: High-quality video generation with cascaded latent diffusion models.
\newblock \emph{IJCV}, 2024{\natexlab{c}}.

\bibitem[Wang et~al.(2004)Wang, Bovik, Sheikh, and Simoncelli]{wang2004image}
Zhou Wang, Alan~C Bovik, Hamid~R Sheikh, and Eero~P Simoncelli.
\newblock Image quality assessment: from error visibility to structural similarity.
\newblock \emph{IEEE TIP}, 2004.

\bibitem[Wang et~al.(2024{\natexlab{d}})Wang, Jiang, Zheng, Wang, He, Wang, Chen, Zhou, et~al.]{wang2024patch}
Zhendong Wang, Yifan Jiang, Huangjie Zheng, Peihao Wang, Pengcheng He, Zhangyang Wang, Weizhu Chen, Mingyuan Zhou, et~al.
\newblock Patch diffusion: Faster and more data-efficient training of diffusion models.
\newblock \emph{NeurIPS}, 2024{\natexlab{d}}.

\bibitem[Weng et~al.(2024)Weng, Feng, Wang, Dai, Wang, Yin, Zhao, Qiu, Bao, Yuan, et~al.]{weng2024art}
Wenming Weng, Ruoyu Feng, Yanhui Wang, Qi Dai, Chunyu Wang, Dacheng Yin, Zhiyuan Zhao, Kai Qiu, Jianmin Bao, Yuhui Yuan, et~al.
\newblock Art-v: Auto-regressive text-to-video generation with diffusion models.
\newblock In \emph{CVPRW}, 2024.

\bibitem[Xie et~al.(2025)Xie, Chen, Chen, Cai, Lin, Zhang, Li, Lu, and Han]{xie2024sana}
Enze Xie, Junsong Chen, Junyu Chen, Han Cai, Yujun Lin, Zhekai Zhang, Muyang Li, Yao Lu, and Song Han.
\newblock Sana: Efficient high-resolution image synthesis with linear diffusion transformers.
\newblock In \emph{ICLR}, 2025.

\bibitem[Xing et~al.(2024{\natexlab{a}})Xing, Xia, Zhang, Chen, Yu, Liu, Liu, Wang, Shan, and Wong]{xing2025dynamicrafter}
Jinbo Xing, Menghan Xia, Yong Zhang, Haoxin Chen, Wangbo Yu, Hanyuan Liu, Gongye Liu, Xintao Wang, Ying Shan, and Tien-Tsin Wong.
\newblock Dynamicrafter: Animating open-domain images with video diffusion priors.
\newblock In \emph{ECCV}, 2024{\natexlab{a}}.

\bibitem[Xing et~al.(2024{\natexlab{b}})Xing, Dai, Hu, Wu, and Jiang]{xing2024simda}
Zhen Xing, Qi Dai, Han Hu, Zuxuan Wu, and Yu-Gang Jiang.
\newblock Simda: Simple diffusion adapter for efficient video generation.
\newblock In \emph{CVPR}, 2024{\natexlab{b}}.

\bibitem[Xing et~al.(2024{\natexlab{c}})Xing, Feng, Chen, Dai, Hu, Xu, Wu, and Jiang]{xing2023survey}
Zhen Xing, Qijun Feng, Haoran Chen, Qi Dai, Han Hu, Hang Xu, Zuxuan Wu, and Yu-Gang Jiang.
\newblock A survey on video diffusion models.
\newblock \emph{ACM Computing Surveys}, 2024{\natexlab{c}}.

\bibitem[Xu et~al.(2016)Xu, Mei, Yao, and Rui]{msrvtt}
Jun Xu, Tao Mei, Ting Yao, and Yong Rui.
\newblock Msr-vtt: A large video description dataset for bridging video and language.
\newblock In \emph{CVPR}, 2016.

\bibitem[Xu et~al.(2024)Xu, Zhao, Xiao, and Hou]{xu2024ufogen}
Yanwu Xu, Yang Zhao, Zhisheng Xiao, and Tingbo Hou.
\newblock Ufogen: You forward once large scale text-to-image generation via diffusion gans.
\newblock In \emph{CVPR}, 2024.

\bibitem[Yang et~al.(2025)Yang, Teng, Zheng, Ding, Huang, Xu, Yang, Hong, Zhang, Feng, et~al.]{yang2024cogvideox}
Zhuoyi Yang, Jiayan Teng, Wendi Zheng, Ming Ding, Shiyu Huang, Jiazheng Xu, Yuanming Yang, Wenyi Hong, Xiaohan Zhang, Guanyu Feng, et~al.
\newblock Cogvideox: Text-to-video diffusion models with an expert transformer.
\newblock In \emph{ICLR}, 2025.

\bibitem[Yin et~al.(2025)Yin, Zhang, Zhang, Freeman, Durand, Shechtman, and Huang]{yin2025slow}
Tianwei Yin, Qiang Zhang, Richard Zhang, William~T Freeman, Fredo Durand, Eli Shechtman, and Xun Huang.
\newblock From slow bidirectional to fast autoregressive video diffusion models.
\newblock In \emph{CVPR}, 2025.

\bibitem[Yu et~al.(2023)Yu, Sohn, Kim, and Shin]{yu2023video}
Sihyun Yu, Kihyuk Sohn, Subin Kim, and Jinwoo Shin.
\newblock Video probabilistic diffusion models in projected latent space.
\newblock In \emph{CVPR}, 2023.

\bibitem[Yu et~al.(2024)Yu, Nie, Huang, Li, Shin, and Anandkumar]{yuefficient}
Sihyun Yu, Weili Nie, De-An Huang, Boyi Li, Jinwoo Shin, and Anima Anandkumar.
\newblock Efficient video diffusion models via content-frame motion-latent decomposition.
\newblock In \emph{ICLR}, 2024.

\bibitem[Zeng et~al.(2024)Zeng, Wei, Zheng, Zou, Wei, Zhang, and Li]{zeng2024make}
Yan Zeng, Guoqiang Wei, Jiani Zheng, Jiaxin Zou, Yang Wei, Yuchen Zhang, and Hang Li.
\newblock Make pixels dance: High-dynamic video generation.
\newblock In \emph{CVPR}, 2024.

\bibitem[Zhang et~al.(2024)Zhang, Wu, Liu, Zhao, Ran, Gu, Gao, and Shou]{zhang2024show}
David~Junhao Zhang, Jay~Zhangjie Wu, Jia-Wei Liu, Rui Zhao, Lingmin Ran, Yuchao Gu, Difei Gao, and Mike~Zheng Shou.
\newblock Show-1: Marrying pixel and latent diffusion models for text-to-video generation.
\newblock \emph{IJCV}, 2024.

\bibitem[Zhang et~al.(2025)Zhang, Wu, Xing, Shao, and Jiang]{zhang2023adadiff}
Hui Zhang, Zuxuan Wu, Zhen Xing, Jie Shao, and Yu-Gang Jiang.
\newblock Adadiff: Adaptive step selection for fast diffusion.
\newblock In \emph{AAAI}, 2025.

\bibitem[Zhang et~al.(2018{\natexlab{a}})Zhang, Isola, Efros, Shechtman, and Wang]{zhang2018perceptual}
Richard Zhang, Phillip Isola, Alexei~A Efros, Eli Shechtman, and Oliver Wang.
\newblock The unreasonable effectiveness of deep features as a perceptual metric.
\newblock In \emph{CVPR}, 2018{\natexlab{a}}.

\bibitem[Zhang et~al.(2018{\natexlab{b}})Zhang, Isola, Efros, Shechtman, and Wang]{zhang2018unreasonable}
Richard Zhang, Phillip Isola, Alexei~A Efros, Eli Shechtman, and Oliver Wang.
\newblock The unreasonable effectiveness of deep features as a perceptual metric.
\newblock In \emph{CVPR}, 2018{\natexlab{b}}.

\bibitem[Zhang et~al.(2023)Zhang, Wang, Zhang, Zhao, Yuan, Qin, Wang, Zhao, and Zhou]{zhang2023i2vgen}
Shiwei Zhang, Jiayu Wang, Yingya Zhang, Kang Zhao, Hangjie Yuan, Zhiwu Qin, Xiang Wang, Deli Zhao, and Jingren Zhou.
\newblock I2vgen-xl: High-quality image-to-video synthesis via cascaded diffusion models.
\newblock \emph{arXiv preprint arXiv:2311.04145}, 2023.

\bibitem[Zhao et~al.(2025)Zhao, Jin, Wang, and You]{zhao2024real}
Xuanlei Zhao, Xiaolong Jin, Kai Wang, and Yang You.
\newblock Real-time video generation with pyramid attention broadcast.
\newblock In \emph{ICLR}, 2025.

\bibitem[Zheng et~al.(2024)Zheng, Peng, Yang, Shen, Li, Liu, Zhou, Li, and You]{zheng2024open}
Zangwei Zheng, Xiangyu Peng, Tianji Yang, Chenhui Shen, Shenggui Li, Hongxin Liu, Yukun Zhou, Tianyi Li, and Yang You.
\newblock Open-sora: Democratizing efficient video production for all.
\newblock \emph{arXiv preprint arXiv:2412.20404}, 2024.

\end{thebibliography}

}

\newpage
\appendix

\section{Implementation details}

\begin{figure}[ht]
\centering
% \vspace{-0.1in}
\includegraphics[width=\linewidth]{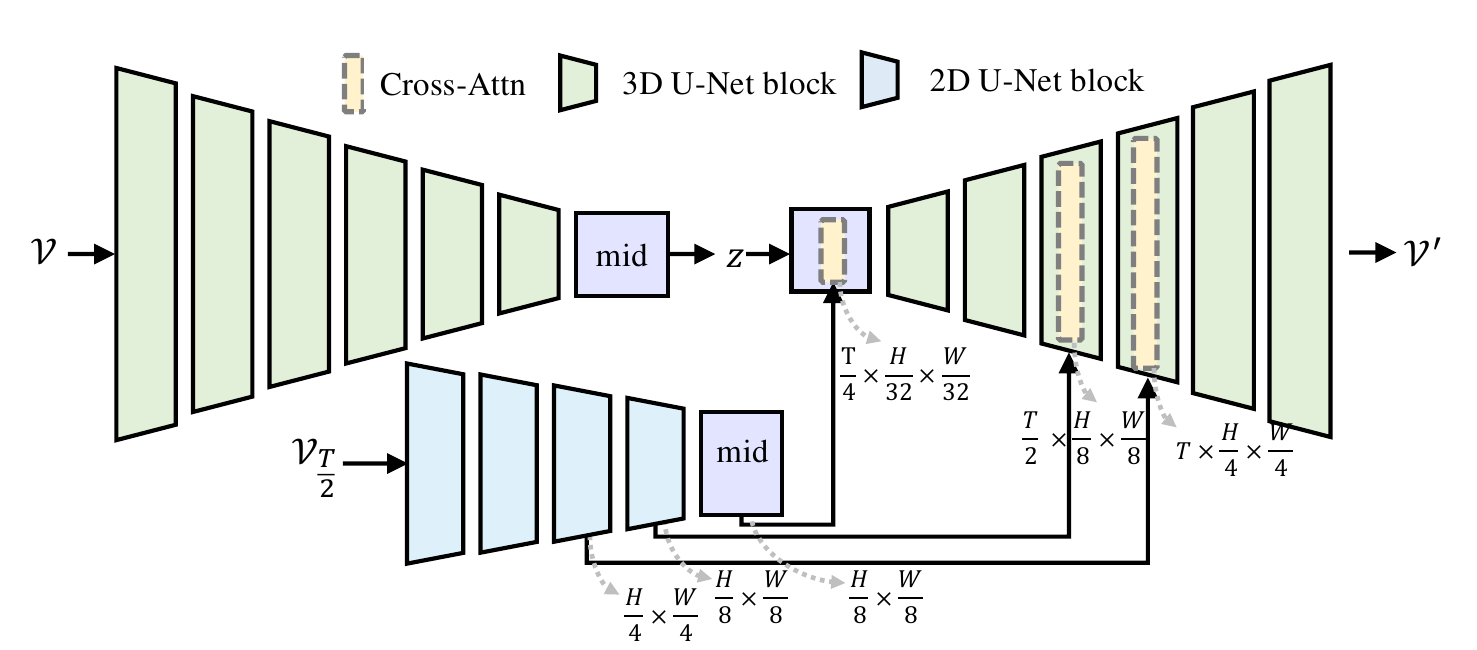}
\vspace{-0.1in}
\caption{The detailed architecture of \system-VAE with $f_t=4, f_s=32$. }
\label{fig:vae_arch_details}
\vspace{-0.1in}
\end{figure}

\noindent \textbf{\system-VAE. } We demonstrate the detailed architecture of \system in \cref{fig:vae_arch_details}. Generally, we follow the training strategies of SD-VAE~\cite{stablediffusion}. We employ a customized version of PatchGAN~\cite{isola2017image} based on 3D convolutions and optimize the model with $L_1$ loss, KL loss, perceptual loss~\cite{zhang2018unreasonable}, and GAN loss. While initializing the 2D encoder and the 3D VAE with SD-VAE pre-trained weights accelerates convergence, we find that freezing the 2D encoder leads to worse performance than training the full parameters. We follow LaMD~\cite{hu2023lamd} to feed the motion latent into a normalization layer to obtain the output of the VAE encoder. Therefore, in the stage of diffusion training, we use a scale factor of 1.0, which is multiplied by the input latent as input into DiT.

\begin{table}[ht]
\small
\centering  
\ra{1.2}
\caption{Hyperparameters for \system-VAE }
 \begin{tabular}{ccc}
 \toprule
 & \multicolumn{2}{c}{\system-VAE}\\
 \midrule
 $z$-shape\footnote{on $256^2$ videos} & $4 \times 8 \times 8 \times 16 $ & $8 \times 8 \times 8 \times 16 $\\
 Channels (3d) &  \multicolumn{2}{c}{128} \\
 Channels (2d) &  \multicolumn{2}{c}{128} \\
 Ch Multiplier (3d) &  \multicolumn{2}{c}{1,2,2,4,4,4}\\
  Ch Multiplier (2d) &  \multicolumn{2}{c}{1,2,2,4,4}\\
  
 Depth &  \multicolumn{2}{c}{2}\\
 Batch size  & 32 & 24 \\
 Learning rate & \multicolumn{2}{c}{4e-5} \\
 Iterations & \multicolumn{2}{c}{1,000,000} \\
 \bottomrule
\end{tabular}
\end{table}

During inference, We split videos with a resolution over $256 \times 256$ into overlapping spatial tiles, we fuse the encoded latent as well as the video output, in a similar manner with Movie Gen~\cite{polyak2024movie}. Note that since \system-VAE employs a spatial down-sampling factor of 32, \system-VAE can only process video inputs whose width and height are divisible by 32. Otherwise,  videos should be padded to meet this requirement before being fed into the VAE.

\begin{figure*}[ht]
\centering
% \vspace{-0.1in}
\includegraphics[width=0.9\linewidth]{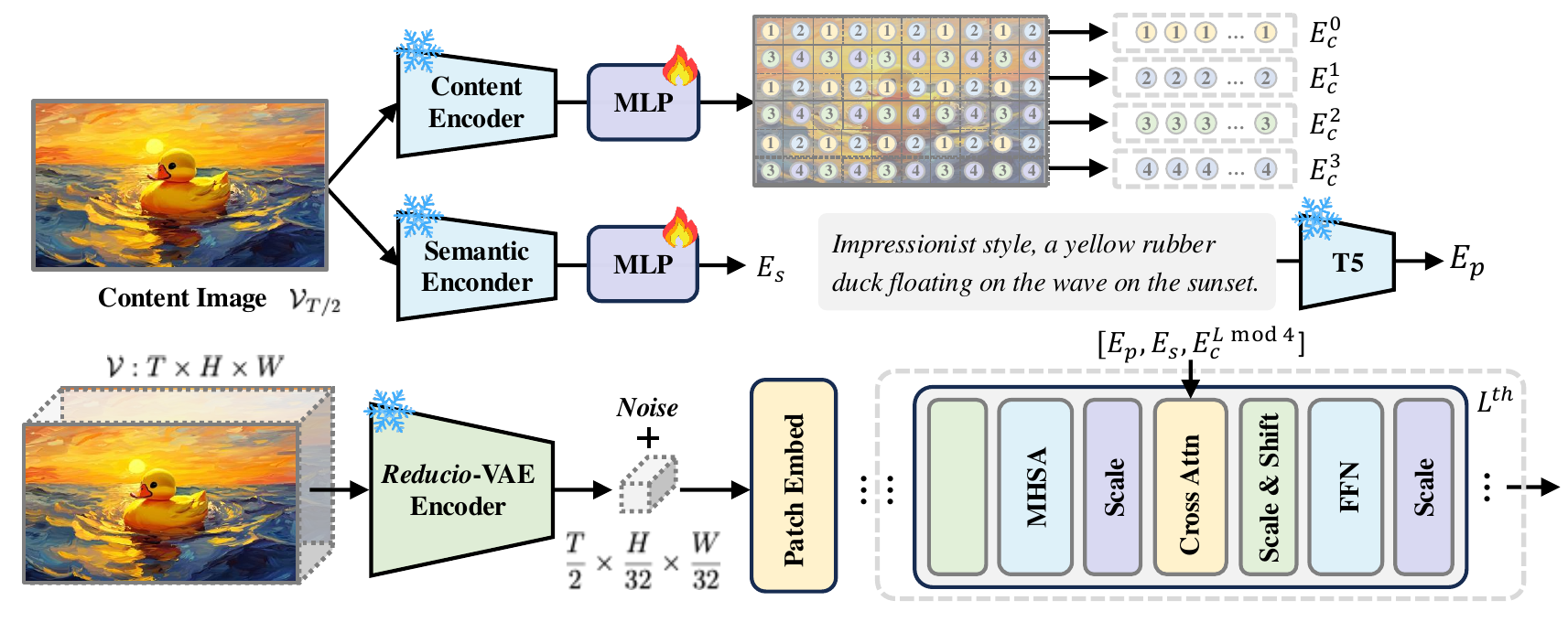}
\vspace{-0.1in}
\caption{The overview of the efficient content condition solution for \system-DiT on high-resolution videos. }
\label{fig:dit_net_highres}
\vspace{-0.1in}
\end{figure*}

\vspace{0.05in}
\noindent \textbf{\system-DiT. } We elaborate on the details of content image conditions in \cref{fig:dit_net_highres}. During inference, we use classifier-free guidance~\cite{ho2022classifier} for better generation quality and set the default scale to 2.5. During training, we randomly drop image conditions at a probability of 0.1, as well as drop 10\% text conditions. 

\section{More ablations}

\begin{table}[h]
  \small
  \centering
    \caption{Ablation on the convolution types in \system-VAE.}
        \vspace{-0.05in}
        \addtolength{\tabcolsep}{8pt}
        \ra{1.2}
         \small
        % \resizebox{0.52\linewidth}{!}
        {\begin{tabular}{cccc}
        \toprule
         {\textbf{$f_s$}} & {\textbf{Conv}} & {\textbf{PSNR$\uparrow$}} & {\textbf{SSIM$\uparrow$}}\\  
        \cmidrule{1-4}
        64 & 2d & 30.22 & 0.86\\ 
       \rowcolor{lightblue} 64 & 3d  & 35.51 & 0.94\\ 
        \bottomrule
        \end{tabular}} 
        \label{tab:vae_conv}
\vspace{-0.1in}
\end{table}

\vspace{0.05in}
\noindent \textbf{Using 3d VAE in \system-VAE} helps to reconstruct videos in a better quality. We keep $f_t$ to 1 and $f_s$ to 32 and implement VAE with 2d convolutions. During decoding, we duplicate the middle frame condition for $T$ times to fuse with the latent of each frame respectively. As shown in \cref{tab:vae_conv}, \system-VAE with 3d convolution outperforms its counterpart with 2d convolution. We believe that 3d convolution facilitates the VAE to model consistent motion and capture spatiotemporal differences.

\begin{table}[h]
  \small
  \centering
    \caption{Ablation on the content frame choice in \system-VAE.}
        \vspace{-0.05in}
        \addtolength{\tabcolsep}{12pt}
        \ra{1.2}
         \small
        {\begin{tabular}{ccc}
        \toprule
        {\textbf{Content Frame}} & {\textbf{PSNR$\uparrow$}} & {\textbf{SSIM$\uparrow$}}\\  
        \cmidrule{1-3}
        n/a & 27.91 & 0.80 \\
        random &31.72 &0.87\\
        \rowcolor{lightblue} middle  &35.88  & 0.94\\
        \bottomrule
        \end{tabular}} 
        \label{tab:vae_cond_frame}
\vspace{-0.1in}
\end{table}

\vspace{0.05in}
\noindent \textbf{Using middle frame in \system-VAE.} The content frame in \system-VAE provides a strong content prior and hence leads to a promising reconstruction performance. On the other hand, relying on any given frame as the content image may not generalize perfectly in all scenarios, especially when certain
entities appear only briefly or outside the chosen frame. As shown in \cref{tab:vae_cond_frame}, we choose the middle frame by default as it serves as a more stable and robust content guidance due to its temporal centrality. Meanwhile, \system-VAE without condition achieves significantly worse results in both PSNR (-7.97) and SSIM (-0.14). In consequence, the \system-DiT framework without condition-based 3D VAE leads to unsatisfactory results featured with blurry frames and obvious visual defects.

\begin{table}[h]
  \small
  \centering
    \caption{Comparison with the state-of-the-art 2D Autoencoder with a significant spatial down-sampling factor.}
        \vspace{-0.05in}
        \addtolength{\tabcolsep}{1pt}
        \ra{1.2}
         \small
    {\begin{tabular}{ccccc}
    \toprule
    {\textbf{Model}}  & {\textbf{latent shape}}& {\textbf{$|z|$}} & {\textbf{PSNR$\uparrow$}} & {\textbf{SSIM$\uparrow$}}\\  
    \midrule
    DC-AE~\cite{chen2024deep} & $16 \times 8\times8$ & 32 &30.68 & 0.70\\
   \rowcolor{lightblue} \system-VAE & $4 \times 8\times8$ & 16& 35.56& 0.97\\
    \bottomrule
    \end{tabular}} 
    \label{tab:cmp_dc_ae}
\vspace{-0.1in}
\end{table}

\vspace{0.05in}
\noindent \textbf{Comparison between DC-AE and \system-VAE.}
We compare \system-VAE with DC-AE~\cite{chen2024deep}  on the Pexel test split with resolution of $512\times 512$. As shown in the Table below, \system-VAE outperforms DC-AE on PSNR and SSIM by 4.88 and 0.27, respectively, highlighting the advantage of our framework in video domain.

\begin{table}[ht]
  \small
  \centering
    \caption{Ablation on the attention type in \system-DiT.}
        \vspace{-0.05in}
        \addtolength{\tabcolsep}{16pt}
        \ra{1.2}
         \small
        % \resizebox{0.52\linewidth}{!}
        {\begin{tabular}{ccc}
        \toprule
        {\textbf{Attn}} & {\textbf{FVD$\downarrow$}} & {\textbf{IS$\uparrow$}}\\  
        \cmidrule{1-3}
        2d + 1d & 382.2 & 32.4 \\
        \rowcolor{lightblue} 3d  & 337.6 & 34.1\\
        \bottomrule
        \end{tabular}} 
        \label{tab:dit_attn}
\vspace{-0.1in}
\end{table}

\begin{table}[ht]
\small
\centering  
\addtolength{\tabcolsep}{4pt}
\ra{1.2}
\caption{Comparison with more SOTA models on Vbench.}
\begin{tabular}{lccc}
 \toprule
\multirow{2}{*}{\textbf{Model}} &\textbf{Quality}   &\textbf{Semantic} & \textbf{Total} \\
& \textbf{Score} &\textbf{Score} &\textbf{Score} \\ 
\midrule
Show-1~\cite{zhang2024show}   & 80.42 &  72.98 & 78.93 \\
Lavie~\cite{wang2023lavie} & 78.78 &  70.31 &  77.08 \\
VideoCrafter~\cite{chen2023videocrafter1}& 81.59 & 72.22 & 79.72\\
OpenSora v1.2~\cite{zheng2024open} &81.35 & 73.39 &79.76\\
Lavie-2~\cite{wang2023lavie} & 83.24 & 75.76 & 81.75\\
Pyramid Flow~\cite{jin2024pyramidal} & 84.74 & 69.62 & 81.72 \\
VideoCrafter-2~\cite{he2024venhancer} & 83.27& 76.73&  81.97\\
\textbf{\system-DiT} & 82.24 & 78.00 & 81.39 \\
WAN~\cite{wan2025wan} &  84.92 & 80.10 & 83.96 \\
STIV~\cite{lin2024stiv}& 81.20 & 72.70 & 79.50 \\
CausVid~\cite{yin2025slow}& 85.21 & 78.57 & 83.88 \\
\bottomrule
\end{tabular} 
\label{tab:vbench_dit_overall}
\vspace{-0.1in}
\ra{1.15}
\end{table}

\begin{table*}[t]
\small
\centering  
\caption{Detailed quantitative comparison with state-of-the-art text-to-video generation models on VBench.}
\vspace{-0.1in}
\addtolength{\tabcolsep}{-3 pt}
\ra{1.2}
 \resizebox{\linewidth}{!}{
 \begin{tabular}{lccccccccccccccccc}
 \toprule
 \multirow{2}{*}{\textbf{Model}}  &{\textbf{subject}}  &  {\textbf{background}} &  {\textbf{temporal}} &\textbf{motion} & \textbf{dynamic}  & \textbf{aesthetic} & \textbf{imaging} & \textbf{object}  & \textbf{multiple} & \textbf{ human} & \multirow{2}{*}{\textbf{color}} & {\textbf{spatial}} &\multirow{2}{*}{\textbf{scene}} &\textbf{appearance} &\textbf{temporal} &\textbf{overall}\\
 
&  {\textbf{consistency}} & {\textbf{consistency}} &{\textbf{flickering}} & \textbf{smoothness} & \textbf{degree} & \textbf{quality} & \textbf{quality} & \textbf{class}  & \textbf{objects}& \textbf{action} &  & {\textbf{relationship}}&  &\textbf{style} &\textbf{style} &\textbf{consistency} \\
 % \cmidrule{1-8}
 \midrule

 Lavie~\cite{wang2023lavie} & 91.41& 97.47& 98.30& 96.38& 49.72& 54.94& 61.90& 91.82& 33.32& 96.80& 86.39& 34.09& 52.69& 23.56& 25.93& 26.41\\
 Show-1~\cite{zhang2024show}  & 95.53 & 98.02&  99.12& 98.24& 44.44& 57.35& 58.66& 93.07& 45.47& 95.60& 86.35& 53.50& 47.03& 23.06& 25.28& 27.46\\
 VideoCrafter~\cite{chen2023videocrafter1}&95.10& 98.04& 98.93& 95.67& 55.00& 62.67& 65.46& 78.18& 45.66& 91.60& 93.32& 58.86& 43.75& 24.41& 25.54& 26.76 \\
 OpenSora v1.2~\cite{zheng2024open} &  96.75 & 97.61 & 99.53 & 98.50 & 42.39 & 56.85 & 63.34 & 82.22 &  51.83 &91.20 & 90.08&  68.56 &  42.44 & 23.95 & 24.54 & 26.85\\
 Lavie-2~\cite{wang2023lavie} & 97.90 & 98.45 & 98.76 & 98.42 & 31.11 &67.62 & 
70.39 & 97.52 & 64.88 & 96.40 & 91.65& 38.68 & 49.59 & 25.09 &  25.24& 27.39 \\
 Pyramid Flow~\cite{jin2024pyramidal} & 96.95& 98.06& 99.49& 99.12& 64.63& 63.26& 65.01& 86.67& 50.71& 85.60& 82.87& 59.53& 43.20& 20.91& 23.09& 26.23 \\
 VideoCrafter-2~\cite{he2024venhancer} & 97.17& 98.54& 98.46& 97.75& 42.50& 65.89& 70.45& 93.39& 49.83& 95.00& 94.41& 64.88& 51.82& 24.32& 25.17& 27.57\\
 \textbf{\system-DiT} & 98.05 & 99.13 & 98.45 & 98.77 & 27.78 & 64.02 &  67.67 & 91.49 & 69.91 & 92.60 & 89.06 & 52.85 & 54.90 & 25.16 & 26.40 & 28.87 \\

\bottomrule
\end{tabular}}
\label{tab:vbench_dit}
% \vspace{-0.1in}
\end{table*}

\begin{figure*}[t]
\centering
\includegraphics[width=\linewidth]{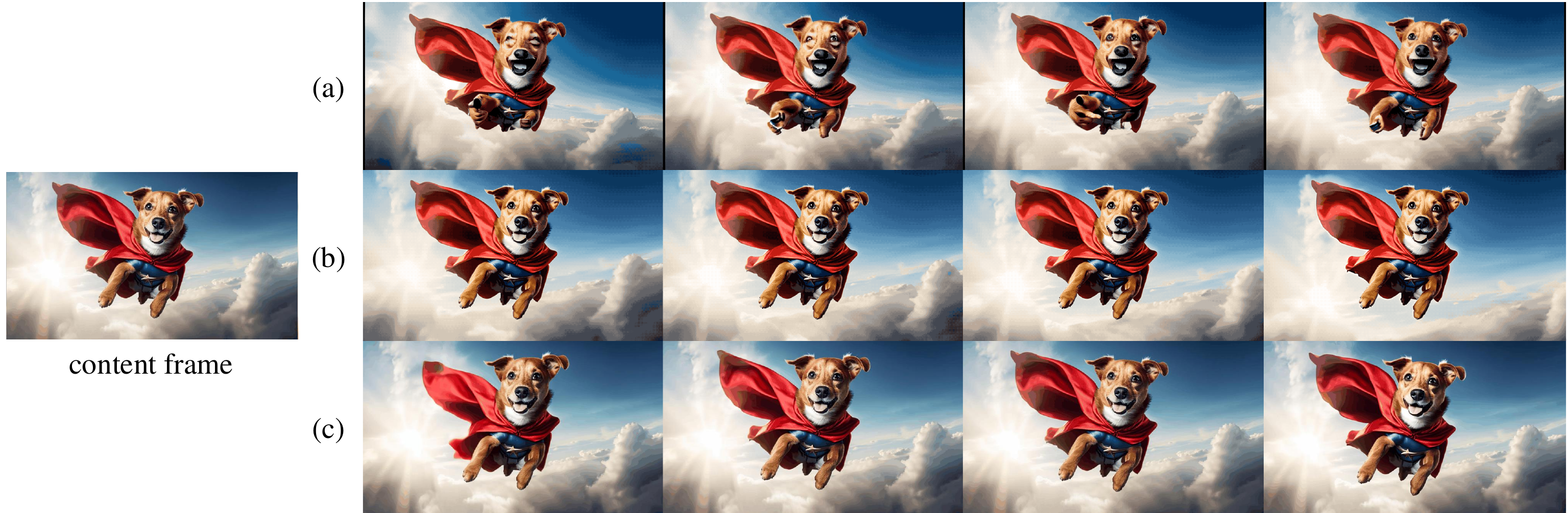}\\
\hspace{0.01in} \textit{\small \blank{4cm} A dog wearing a Superhero outfit with red cape flying through the sky.}

\vspace{0.05in}

\includegraphics[width=\linewidth]{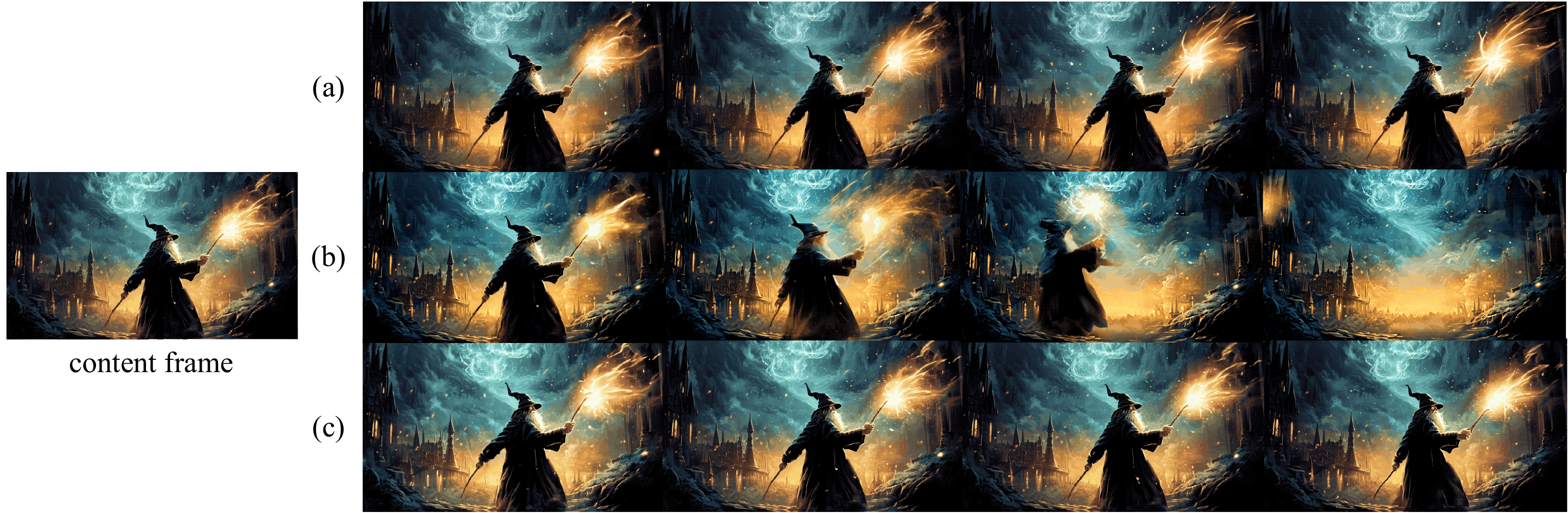}\\
\hspace{0.01in} \textit{\small \blank{4cm} A dog wearing a Superhero outfit with red cape flying through the sky.}
\caption{Comparison between frames generated given an identical frame and prompt, by (a) DynamicCrafter~\cite{xing2025dynamicrafter}, (b) SVD-XT~\cite{blattmann2023stable} and (c) \system-DiT, respectively. We resize the output frames from $1344 \times 768$ to $1024\times576$ to match with the former two baselines.}
\label{fig:i2v_cmpv2}
\vspace{-0.1in}
\end{figure*}

\vspace{0.05in}
\noindent \textbf{Using joint spatiotemporal 3D attention in \system-DiT} outweighs using factorized spatial and temporal attention (\ie, 2D + 1D attention) in generation quality. 
Interestingly, we observe that factorized attention leads to a faster convergence of training loss. However, with the same training step, as shown in \cref{tab:dit_attn}, factorized attention lags behind its counterpart with joint 3D attention for 45 in FVD.
We suppose the possible reason is that 2D + 1D scheme demands adding additional temporal layers and performs factorized self-attention on a small set of tokens e ach, making it hard to model smooth open-set motion with the light computation. In contrast, 3D attention directly exploits the original parameters and collaborates all spatiotemporal tokens.

\vspace{0.05in}
\noindent \textbf{Quantitative results.} We display the detailed performance comparison on VBench~\cite{huang2024vbench} in \cref{tab:vbench_dit} and \cref{tab:vbench_dit_overall}. Despite using only 3.2K A100 GPU hours and 5.4M training samples, \system-DiT achieves a promising semantic score of 78.00, beating a range of state-of-the-art LDMs.  
While the most recent models such as WAN~\cite{wan2025wan} and CausVid~\cite{yin2025slow} achieve higher overall scores than \system-DiT, we argue that our model uses a much smaller scale of training data and has a relatively small model scale, \ie, 1.2B.

\vspace{0.05in}
\noindent \textbf{Visualizations.} 
We present more examples of comparison between \system, SVD-XT~\cite{blattmann2023stable} and DynamicCrafter~\cite{xing2025dynamicrafter} in \cref{fig:i2v_cmpv2}. \system-DiT exhibits reasonable motion and preserves the details in the content frame well.

\end{document}